\documentclass[twocolumn]{asme_mtr2}

\newcommand{\gbf}[1] {\mbox{\boldmath${#1}$\unboldmath}}
\newcommand{\be}{\begin{equation}}
\newcommand{\ee}{\end{equation}}
\newcommand{\beq}{\begin{equation}}
\newcommand{\eeq}{\end{equation}}
\newcommand{\bed}{\begin{displaymath}}
\newcommand{\eed}{\end{displaymath}}
\newcommand{\beqa}{\begin{eqnarray}}
\newcommand{\eeqa}{\end{eqnarray}}
\newcommand{\beqann}{\begin{eqnarray*}}
\newcommand{\eeqann}{\end{eqnarray*}}
\newcommand{\bseq}{\begin{subequations}}
\newcommand{\eseq}{\end{subequations}}

\newcommand{\mat}[2]{\left[ \begin{array}{#1} #2 \end{array} \right] }

\newcommand{\ba}{\begin{array}}
\newcommand{\ea}{\end{array}}

\usepackage{amsbsy}
\usepackage[dvips]{graphicx}
\usepackage{epsf}
\usepackage{epsfig}
\usepackage{array}
\usepackage{amssymb}
\usepackage{amsmath}
\usepackage{subfigure}
\usepackage{subeqn}

\def\confshortname{}
\def\conffullname{}
\def\confdate{}
\def\confmonth{}
\def\confyear{}
\def\confcity{}
\def\confstate{}
\def\confcountry{}

\def\papernum{}

\newcommand{\field}[1]{\mathbb{#1}}

\title{SENSITIVITY ANALYSIS OF THE ORTHOGLIDE,\goodbreak A 3-DOF TRANSLATIONAL PARALLEL KINEMATIC MACHINE}

\author{St\'ephane Caro, Philippe Wenger, Fouad Bennis, Damien Chablat
    \affiliation{
      Institut de Recherche en Communications et Cybern\'etique de Nantes
      \thanks{IRCCyN: UMR n$^\circ$ 6597 CNRS, \'Ecole Centrale de Nantes,
                        Universit\'e de Nantes, \'Ecole des Mines de Nantes} \\
      1, rue de la No\"e - 44321 Nantes, France \\
      Stephane.Caro@irccyn.ec-nantes.fr
    }
}

\begin{document}
\maketitle

\begin{abstract}

In this paper, two complementary methods are introduced to analyze
the sensitivity of a three degree-of-freedom (DOF) translational
Parallel Kinematic Machine (PKM) with orthogonal linear joints:
the Orthoglide. Although these methods are applied to a particular
PKM, they can be readily applied to 3-DOF Delta-Linear PKM such as
ones with their linear joints parallel instead of orthogonal. On
the one hand, a linkage kinematic analysis method is proposed to
have a rough idea of the influence of the length variations of the
manipulator on the location of its end-effector. On the other
hand, a differential vector method is used to study the influence
of the length and angular variations in the parts of the
manipulator on the position and orientation of its end-effector.
Besides, this method takes into account the variations in the
parallelograms. It turns out that variations in the design
parameters of the same type from one leg to another have the same
effect on the position of the end-effector. Moreover, the
sensitivity of its pose to geometric variations is a minimum in
the kinematic isotropic configuration of the manipulator. On the
contrary, this sensitivity approaches its maximum close to the
kinematic singular configurations of the manipulator.

\end{abstract}

\begin{nomenclature}
\begin{description}
  \item[${\cal R}_b (O,x,y,z)$]: reference coordinate frame centered at $O$, the intersection between the directions of the three actuated prismatic joints.
  \item[${\cal R}_P (P,X,Y,Z)$]: coordinate frame attached to the end-effector.
  \item[${\cal R}_i (A_i,x_i,y_i,z_i)$]: coordinate frame attached
  to the $i^{th}$ prismatic joint, $i = 1,2,3$.
  \item[${\bf p} = \mat{ccc}{p_x & p_y & p_z}^{T} $]: vector of the Cartesian coordinates of the end-effector,
  expressed in ${\cal R}_b$.
  \item[$\delta {\bf p} = \mat{ccc}{\delta p_x & \delta p_y & \delta p_z}^{T} $]: position error of the end-effector,
  expressed in ${\cal R}_b$.
  \item[$\delta \theta = \mat{ccc}{\delta \theta_x & \delta \theta_y & \delta \theta_z}^{T}$]: orientation error of the end-effector,
  expressed in ${\cal R}_b$.
  \item[$\rho_i$]: displacement of the $i^{th}$ prismatic joint.
  \item[$\delta \rho_i$]: displacement error of the $i^{th}$ prismatic joint.
  \item[$L_i$]: theoretical length of the $i^{th}$ parallelogram.
  \item[$A_i, B_i, C_i$]: depicted in Fig. \ref{fig:orthoglide}.
  \item[$a_i$]: distance between points $O$ and $A_i$.
  \item[$r_i$]: distance between points $P$ and $C_i$.
  \item[$b_{1y}, b_{1z}$]: position errors of point $B_1$ along $y$ and $z$ axes, respectively.
  \item[$b_{2x}, b_{2z}$]: position errors of point $B_2$ along $x$ and $z$ axes, respectively.
  \item[$b_{3x}, b_{3y}$]: position errors of point $B_3$ along $x$ and $y$ axes, respectively.
  \item[$h_{1} = b_{1y}, \ k_{1} = b_{1z}, \ h_{2} = b_{2x}, \ k_{2}=b_{2z}, \ h_{3} = b_{3x}, \ k_{3} = b_{3y}.$]
  \item[$d_i$]: nominal width of the $i^{th}$ parallelogram.
  \item[$\delta L_i$]: variation in the length of the $i^{th}$ parallelogram.
  \item[$\delta L_{ij}$]: variation in the length of link $\overline{B_{ij}C_{ij}}$, $j = 1,2$ (see Fig. \ref{fig:schema-cinematique-jambe}).
  \item[$\delta b_i$]: variation in the length of link $\overline{B_{i1}B_{i2}}$.
  \item[$\delta c_i$]: variation in the length of link $\overline{C_{i1}C_{i2}}$.
  \item[$\delta l_i$]: parallelism error of links $\overline{B_{i1}B_{i2}}$ and $\overline{C_{i1}C_{i2}}$.
  \item[$\delta m_i$]: parallelism error of links $\overline{B_{i1}C_{i1}}$ and $\overline{B_{i2}C_{i2}}$.
  \item[${\bf w}_i$]: direction of links $\overline{B_{i1}C_{i1}}$ and $\overline{B_{i2}C_{i2}}$.
  \item[$\delta {\bf w}_i$]: variation in the direction of links $\overline{B_{i1}C_{i1}}$ and $\overline{B_{i2}C_{i2}}$.
  \item[$\delta {\bf e}_i$]: sum of the position errors of points $A_i$, $B_i$, $C_i$.
  \item[$\delta \boldsymbol{\theta}_{Ai}= \mat{ccc}{\delta \theta_{Aix} & \delta \theta_{Aiy} & \delta \theta_{Aiz}}^{T}$]: angular variation in the direction of the $i^{th}$ prismatic joint.
  \item[$\delta \theta_{Bi}= \mat{ccc}{\delta \theta_{Bix} & \delta \theta_{Biy} & \delta \theta_{Biz}}^{T}$]: angular variation between $\overline{B_{i1}B_{i2}}$ and the direction of the $i^{th}$ prismatic joint.
  \item[$\delta \theta_{Ci}= \mat{ccc}{\delta \theta_{Cix} & \delta \theta_{Ciy} & \delta \theta_{Ciz}}^{T}$]: angular variation between the end-effector and $\overline{C_{i1}C_{i2}}$.
  \item[$\delta {\gamma}_i= \mat{ccc}{\delta {\gamma}_{ix} & \delta {\gamma}_{iy} & \delta {\gamma}_{iz}}^{T}$] sum of the orientation errors of the $i^{th}$ parallelogram with respect to the $i^{th}$ prismatic joint and the end-effector.
  \item[\rm{DOF}]: degree-of-freedom.
  \item[\rm{PKM}]: parallel kinematic machine.
\end{description}


\end{nomenclature}

\section{Introduction}
For two decades, parallel manipulators have attracted the
attention of more and more researchers who consider them as
valuable alternative design for robotic mechanisms. As stated by
numerous authors, conventional serial kinematic machines have
already reached their dynamic performance limits, which are
bounded by high stiffness of the machine components required to
support sequential joints, links and actuators. Thus, while having
good operating characteristics (large workspace, high flexibility
and manoeuvrability), serial manipulators have disadvantages of
low stiffness and low power. Conversely, parallel kinematic
machines (PKM) offer essential advantages over their serial
counterparts (lower moving masses, higher stiffness and
payload-to-weight ratio, higher natural frequencies, better
accuracy, simpler modular mechanical construction, possibility to
locate actuators on the fixed base).

However, PKM are not necessarily more accurate than their serial
counterparts. Indeed, even if the dimensional variations can be
compensated with PKM, they can also be amplified contrary to with
their serial counterparts, \cite{wenger99}. Wang et al.
\cite{wang93} studied the effect of manufacturing tolerances on
the accuracy of a Stewart platform. Kim et al. \cite{kim00} used a
forward error bound analysis to find the error bound of the
end-effector of a Stewart platform when the error bounds of the
joints are given, and an inverse error bound analysis to determine
those of the joints for the given error bound of the end-effector.
Kim and Tsai \cite{kim03} studied the effect of misalignment of
linear actuators of a 3-DOF translational parallel manipulator on
the motion of its moving platform. Han et al. \cite{han02} used a
kinematic sensitivity analysis method to explain the gross motions
of a 3-UPU parallel mechanism, and they showed that it is highly
sensitive to certain minute clearances. Fan et al. \cite{fan03}
analyzed the sensitivity of the 3-PRS parallel kinematic spindle
platform of a serial-parallel machine tool. Verner et
al.~\cite{VERNER05} presented a new method for optimal calibration
of PKM based on the exploitation of the least error sensitive
regions in their workspace and geometric parameters space. As a
matter of fact, they used a Monte Carlo simulation to determine
and map the sensitivities to geometric parameters. Moreover, Caro
et al.~\cite{CARO05} developed a tolerance synthesis method for
mechanisms based on a robust design approach.

This paper aims at analyzing the sensitivity of the Orthoglide to
its dimensional and angular variations. The Orthoglide is a three
degree-of-freedom (DOF) translational PKM developed by Chablat and
Wenger \cite{CHABLAT03}. A small-scale prototype of this
manipulator was built at IRCCyN.

Here, the sensitivity of the Orthoglide is studied by means of two
complementary methods. First, a linkage kinematic analysis is used
to have a rough idea of the influence of the dimensional
variations to its end-effector and to show that the variations in
design parameters of the same type from one leg to another have
the same influence on the location of the end-effector. Although
this method is compact, it cannot be used to know the influence of
the variations in the parallelograms. Thus, a differential vector
method is developed to study the influence of the dimensional and
angular variations in the parts of the manipulator, and
particularly variations in the parallelograms, on the position and
the orientation of its end-effector.

In the isotropic kinematic configuration, the end-effector of the
manipulator is located at the intersection between the directions
of its three actuated prismatic joints, and the condition number
of its kinematic Jacobian matrix is equal to one, \cite{zanganeh}.
It is shown that this configuration is the least sensitive one to
geometrical variations, contrary to the closest configurations to
its kinematic singular configurations, which are the most
sensitive to geometrical variations.

Although the two sensitivity analysis methods are applied to a
particular PKM, these methods can be readily applied to other
3-DOF Delta-linear PKM such as ones with parallel linear joints
instead of orthogonal ones.

\section{Manipulator Geometry}
\label{section:manipulator-geometry}

\begin{figure}[!htbp]
\centering\includegraphics[width=80mm]{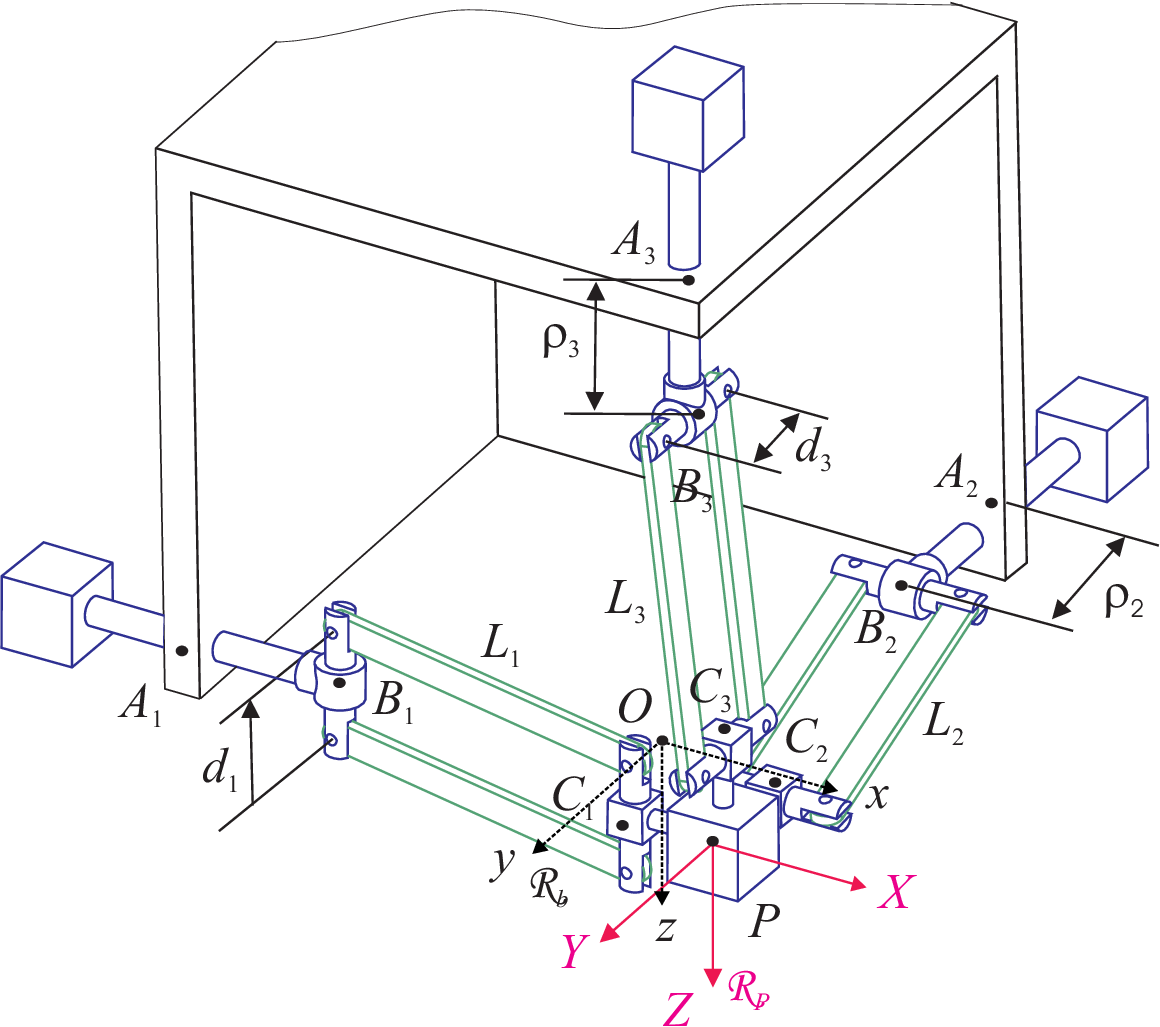}
\caption{Basic kinematic architecture of the Orthoglide}
\label{fig:orthoglide}
\end{figure}

\begin{figure}[!htbp]
\centering\includegraphics[width=80mm]{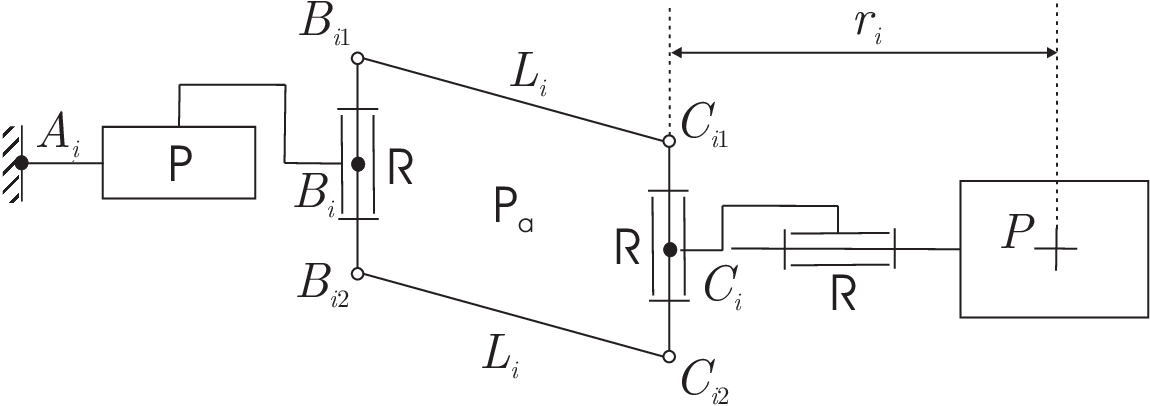}
\caption{Morphology of the $i^{th}$ leg of the Orthoglide}
\label{fig:schema-cinematique-jambe}
\end{figure}

The kinematic architecture of the Orthoglide is shown in
Fig.\ref{fig:orthoglide}. It consists of three identical parallel
chains that are formally described as PR$\mbox{P}_a$RR, where P, R
and $\mbox{P}_a$ denote the prismatic, revolute, and parallelogram
joints respectively, as shown in
Fig.\ref{fig:schema-cinematique-jambe}.

The mechanism input is made up of three actuated orthogonal
prismatic joints. The output body (with a tool mounting flange) is
connected to the prismatic joints through a set of three kinematic
chains. Inside each chain, one parallelogram is used and oriented
in a manner that the output body is restricted to translational
movements only.

The small-scale prototype of the Orthoglide was designed to reach
Cartesian velocity of 1.2 m/s and an acceleration of 17 m/s$^2$.
The desired payload is 4 kg (spindle, tool, included). The size of
its prescribed cubic workspace, $C_u$, is $200 \times 200 \times
200$ mm, where the velocity transmission factors are bounded
between $1/2$ and $2$. The three legs are supposed to be
identical. According to \cite{CHABLAT03}, the nominal lengths,
$L_i$, and widths, $d_i$, of the parallelograms, and the nominal
distances, $r_i$, between points $C_i$ and the end-effector $P$
are identical, {\it i.e.}: $L = L_1 = L_2 = L_3 = 310.58$ mm, $d =
d_1 = d_2 = d_3 = 80$ mm, $r = r_1 = r_2 = r_3 = 31$ mm.

\begin{figure}[!htbp]
\centering\includegraphics[width=80mm]{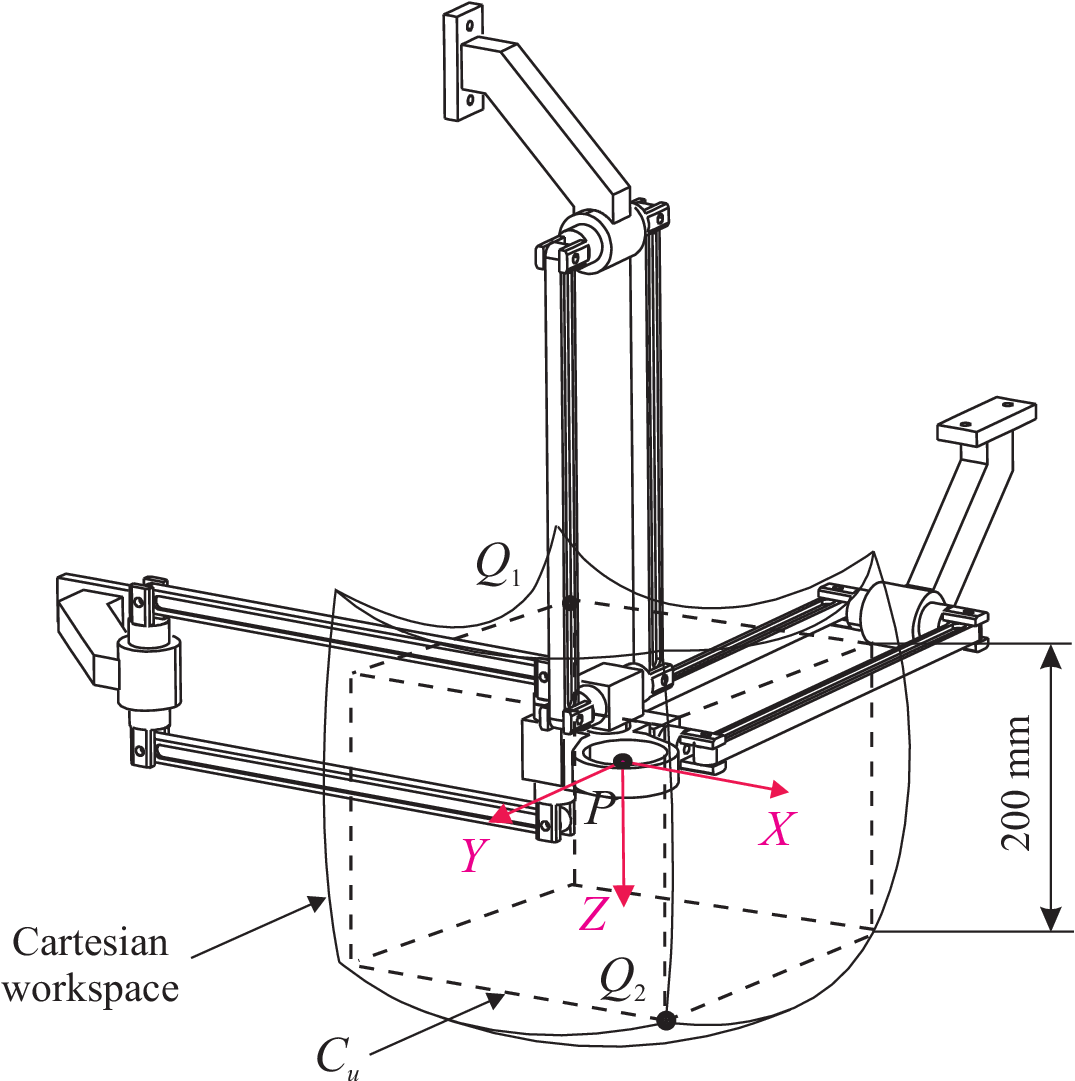}
\caption{Cartesian workspace, $C_u$, points $Q_1$ and $Q_2$}
\label{fig:Orthoglide-Workspace-Q1Q2}
\end{figure}

As depicted in Fig.\ref{fig:Orthoglide-Workspace-Q1Q2}, $Q_1$ and
$Q_2$, vertices of $C_u$, are defined at the intersection between
the Cartesian workspace boundary and the axis $x=y=z$ expressed in
the reference coordinate frame ${\cal R}_b$. $Q_1$ and $Q_2$ are
the closest points to the singularity surfaces. Their Cartesian
coordinates, expressed in ${\cal R}_b$, are equal to
(-73.21,-73.21,-73.21) and (126.79,126.79,126.79), respectively.

The parts of the manipulator are supposed to be rigid-bodies and
there is no joint clearance. The legs of the manipulator, composed
of one prismatic joint, one parallelogram, and three revolute
joints, generate a five DOF motion each. Besides, they are
identical. Therefore, according to Karouia et al. \cite{mourad},
the manipulator is isostatic. Thus, the results obtained by the
sensitivity analysis methods developed in this paper are
meaningful.


\section{Sensitivity Analysis}
\label{chapitre:sensitivity-Fan}
Two complementary methods are used to study the sensitivity of the
Orthoglide. First, a linkage kinematic analysis is used to have a
rough idea of the influence of the dimensional variations to its
end-effector. Although this method is compact, it cannot be used
to know the influence of the variations in the parallelograms.
Thus, a differential vector method is used to study the influence
of the dimensional and angular variations in the parts of the
manipulator, and particularly variations in the parallelograms, on
the position and the orientation of its end-effector.

\subsection{Linkage Kinematic Analysis} \label{section:linkage-kinematic-analysis}
This method aims at computing the sensitivity coefficients of the
position of the end-effector, $P$, to the design parameters of the
manipulator. First, three implicit functions depicting the
kinematic of the manipulator are obtained. A relation between the
variations in the position of $P$ and the variations in the design
parameters follows from these functions. Finally a sensitivity
matrix, which gathers the sensitivity coefficients of $P$, follows
from the previous relation written in matrix form.

\subsubsection{Formulation}
\label{section:linkage-kinematic-analysis-theory}
Figure \ref{fig:orthoglide} depicts the design parameters taken
into account. Points $A_1$, $A_2$, and $A_3$ are the bases of the
prismatic joints. Their Cartesian coordinates, expressed in ${\cal
R}_b$, are ${\bf a}_1$, ${\bf a}_2$, and ${\bf a}_3$,
respectively.
\begin{subequations}\label{E:Cartesian-coord-A1A2A3}
    \begin{gather}
        {\bf a}_1 = \mat{ccc}{-a_1 & 0 & 0}^{T} \label{E:Cartesian-coord-A1} \\
        {\bf a}_2 = \mat{ccc}{0 & -a_2 & 0}^{T} \label{E:Cartesian-coord-A2} \\
        {\bf a}_3 = \mat{ccc}{0 & 0 & -a_3}^{T} \label{E:Cartesian-coord-A3}
    \end{gather}
\end{subequations}
\noindent where $a_i$ is the distance between points $A_i$ and
$O$, the origin of ${\cal R}_b$. Points $B_1$, $B_2$ and $B_3$ are
the links between the prismatic and parallelogram joints. Their
Cartesian coordinates, expressed in ${\cal R}_b$ are:
\begin{subequations}\label{E:Cartesian-coord-B1B2B3}
    \begin{gather}
        {\bf b}_1 = \mat{c}{-a_1 + \rho_1 \\ b_{1y} \\ b_{1z}} \label{E:Cartesian-coord-B1} \\
        {\bf b}_2 = \mat{c}{ b_{2x} \\ -a_2 + \rho_2 \\ b_{2z}} \label{E:Cartesian-coord-B2} \\
        {\bf b}_3 = \mat{c}{ b_{3x} \\ b_{3y} \\ -a_3 + \rho_3 } \label{E:Cartesian-coord-B3}
    \end{gather}
\end{subequations}
\noindent where $\rho_i$ is the displacement of the $i^{th}$
prismatic joint. $b_{1y}$ and $b_{1z}$ are the position errors of
point $B_1$ according to $y$ and $z$ axes. $b_{2x}$ and $b_{2z}$
are the position errors of point $B_2$ according to $x$ and $z$
axes. $b_{3x}$ and $b_{3y}$ are the position errors of point $B_3$
according to $x$ and $y$ axes. These errors result from the
orientation errors of the directions of the prismatic actuated
joints. The Cartesian coordinates of $C_1$, $C_2$, and $C_ 3$,
expressed in ${\cal R}_b$, are the following:
\begin{subequations}\label{E:Cartesian-coord-C1C2C3}
    \begin{gather}
        {\bf c}_1 = \mat{ccc}{p_x-r_1 & 0 & 0}^{T} \label{E:Cartesian-coord-C1} \\
        {\bf c}_2 = \mat{ccc}{0 & p_y-r_2 & 0}^{T} \label{E:Cartesian-coord-C2} \\
        {\bf c}_3 = \mat{ccc}{0 & 0 & p_z-r_3}^{T} \label{E:Cartesian-coord-C3}
    \end{gather}
\end{subequations}
\noindent where ${\bf p} = {\mat{ccc}{p_x & p_y & p_z}}^{T}$ is
the vector of the Cartesian coordinates of the end-effector $P$,
expressed in ${\cal R}_b$.

The expressions of the nominal lengths of the parallelograms
follow from eq.(\ref{eq:Li}),
\begin{equation}\label{eq:Li}
  L_i  = \| {\bf c}_i - {\bf b}_i \|_2 \ , \ i = 1,2,3
\end{equation}

\noindent where $L_i$ is the nominal length of the $i^{th}$
parallelogram and $\|.\|_2$ is the Euclidean norm. Three implicit
functions follow from eq.(\ref{eq:Li}) and are given by the
following equations:
\begin{eqnarray*}\label{eq:fonctions-implicites}
    F_1 & = & (-r_1 + p_x + a_1 - \rho_1)^2 + (p_y - b_{1y})^2+(p_z -
    b_{1z})^2 - L_1^2 = 0 \\
    F_2 & = & (p_x - b_{2x})^2+( -r_2 + p_y + a_2 - \rho_2)^2+(p_z -
    b_{2z})^2 - L_2^2 = 0 \\
    F_3 & = & (p_x - b_{3x})^2+(p_y - b_{3y})^2+(-r_3 + p_z + a_3 -
     \rho_3)^2 - L_3^2 = 0
\end{eqnarray*}
By differentiating functions $F_1$, $F_2$, and $F_3$, with respect
to the design parameters of the manipulator and the position of
the end-effector, we obtain a relation between the positioning
error of the end-effector, $\delta {\bf p}$, and the variations in
the design parameters, $\delta {\bf q}_i$.
\begin{equation}\label{eq:differenciation-fct-implicite}
  {\delta F}_i = {\bf A}_i \delta {\bf p} + {\bf B}_i \delta {\bf
  q}_i = 0 \ , \ i = 1,2,3
\end{equation}
\noindent with
\begin{eqnarray}\label{eq:matrices-expression}
{\bf A}_i &=& \mat{ccc}{{\partial F_i}/{\partial p_x} & {\partial
F_i}/{\partial p_y} & {\partial F_i}/{\partial p_z}} \\
{\bf B}_i &=& \mat{cccccc}{ {\partial F_i}/{\partial a_i} &
{\partial F_i}/{\partial b_{iy}} & {\partial F_i}/{\partial
b_{iz}} & {\partial F_i}/{\partial \rho_i} & {\partial
F_i}/{\partial L_i} & {\partial F_i}/{\partial r_i} } \\
\delta {\bf p} &=& {\mat{ccc}{\delta p_x & \delta p_y & \delta
p_z}}^{T} \\
\delta {\bf q}_i &=& \mat{cccccc}{ {\delta a_i} & {\delta h_{i}} &
{\delta k_{i}} & {\delta \rho_i} & {\delta L_i} & {\delta r_i} }^T
\end{eqnarray}

\noindent where ${\delta a_i}$, ${\delta h_{i}}$, ${\delta
k_{i}}$, ${\delta \rho_i}$, ${\delta L_i}$, and ${\delta r_i}$,
depict the variations in $a_i$, $h_{i}$, $k_{i}$, $\rho_i$, $L_i$,
and $r_i$, respectively with $h_{1} = b_{1y}$, $k_{1} = b_{1z}$,
$h_{2} = b_{2x}$, $k_{2} = b_{2z}$, $h_{3} = b_{3x}$, $k_{3} =
b_{3y}$.

Integrating the three loops of
eq.(\ref{eq:differenciation-fct-implicite}) together and
separating the position parameters and design parameters to
different sides yields the following simplified matrix form:
\begin{equation}\label{eq:forme-matricielle}
  {\bf A} \delta {\bf p} + {\bf B} \delta {\bf q} = 0
\end{equation}

\noindent with
\begin{eqnarray}\label{eq:matrices-expression}
  {\bf A} & = & {\mat{ccc}{{\bf A}_1^T {\bf A}_2^T {\bf A}_3^T}}^T \in {\field{R}}^{3 \times 3} \\
  {\bf B} & = & \mat{ccc}{ {\bf B}_1 & 0 & 0 \\
  0 & {\bf B}_2 & 0 \\
  0 & 0 & {\bf B}_3} \in {\field{R}}^{3 \times 18} \\
  \delta {\bf q} & = & \mat{ccc}{ \delta {\bf q}_1^T & \delta {\bf q}_2^T & \delta {\bf q}_3^T }^T \in {\field{R}}^{18 \times 1}
\end{eqnarray}

Equation (\ref{eq:forme-matricielle}) takes into account the
coupling effect of the three independent structure loops.
According to \cite{CHABLAT03}, {\bf A} is the parallel Jacobian
kinematic matrix of the Orthoglide, which does not meet parallel
kinematic singularities when its end-effector covers $C_u$.
Therefore, {\bf A} is not singular and its inverse, ${\bf
A}^{-1}$, exists. Thus, the positioning error of the end-effector
can be computed using eq.(\ref{eq:matriceC}).
\begin{equation}\label{eq:matriceC}
   \delta {\bf p} = {\bf C} \ \delta {\bf q}
\end{equation}

\noindent where
\begin{equation}\label{eq:matriceC-2}
  {\bf C} = -{\bf A}^{-1} {\bf B} = \mat{ccccc}{\partial p_x / \partial a_1 & \partial p_x / \partial h_{1}  & \cdots & \partial p_x / \partial r_{3} \\
  \partial p_y / \partial a_1 & \partial p_y / \partial h_{1} & \cdots & \partial p_y / \partial r_{3}\\
  \partial p_z / \partial a_1 & \partial p_z / \partial h_{1} & \cdots & \partial p_z / \partial r_{3}
  } \in {\field{R}}^{3 \times 18}
\end{equation}

\noindent represents the sensitivity matrix of the manipulator.
The terms of ${\bf C}$ are the sensitivity coefficients of the
Cartesian coordinates of the end-effector to the design parameters
and are used to analyze the sensitivity of the Orthoglide.

\subsubsection{Results of the Linkage Kinematic Analysis}
\label{section:linkage-kinematic-analysis-results}
The sensitivity matrix ${\bf C}$ of the manipulator depends on the
position of its end-effector.

\begin{figure}[!htbp]
\centering\includegraphics[width=60mm]{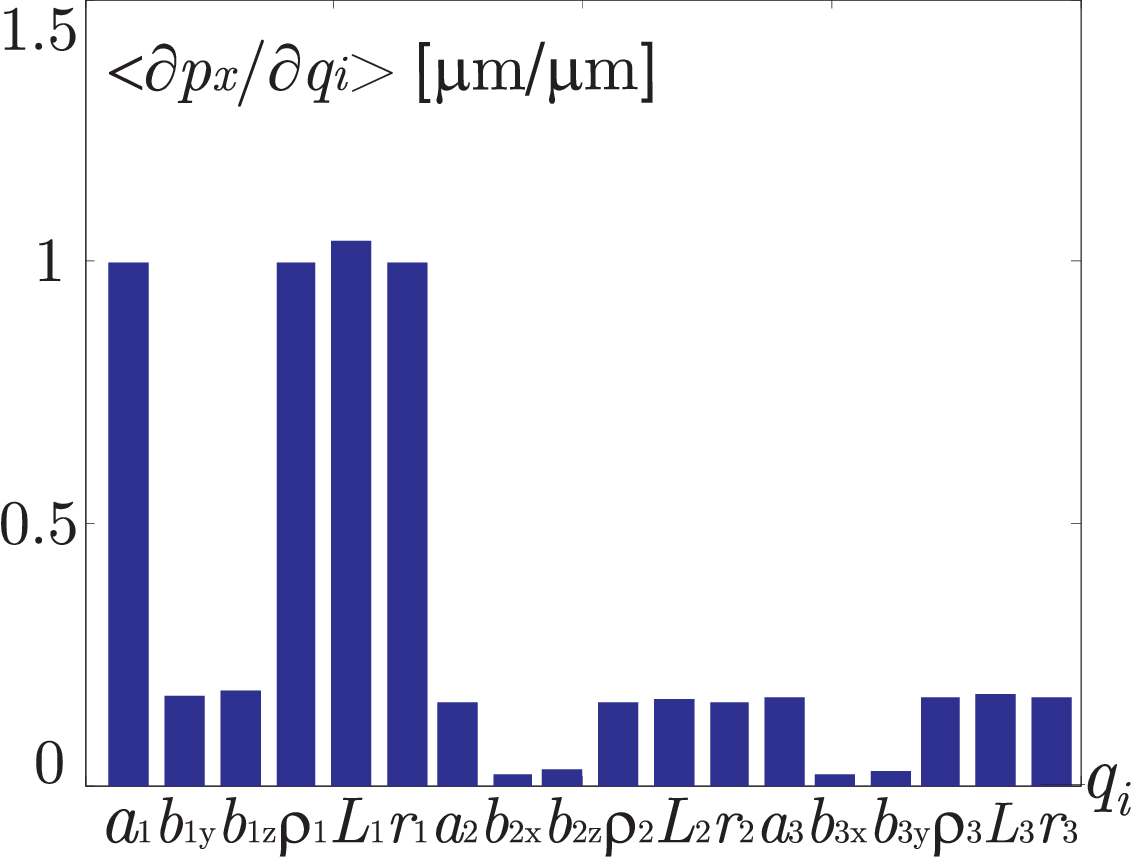}
\caption{Mean of sensitivity of $p_x$ throughout $C_u$}
\label{fig:sensi-px-Cu}
\end{figure}

\begin{figure}[!htbp]
\centering\includegraphics[width=60mm]{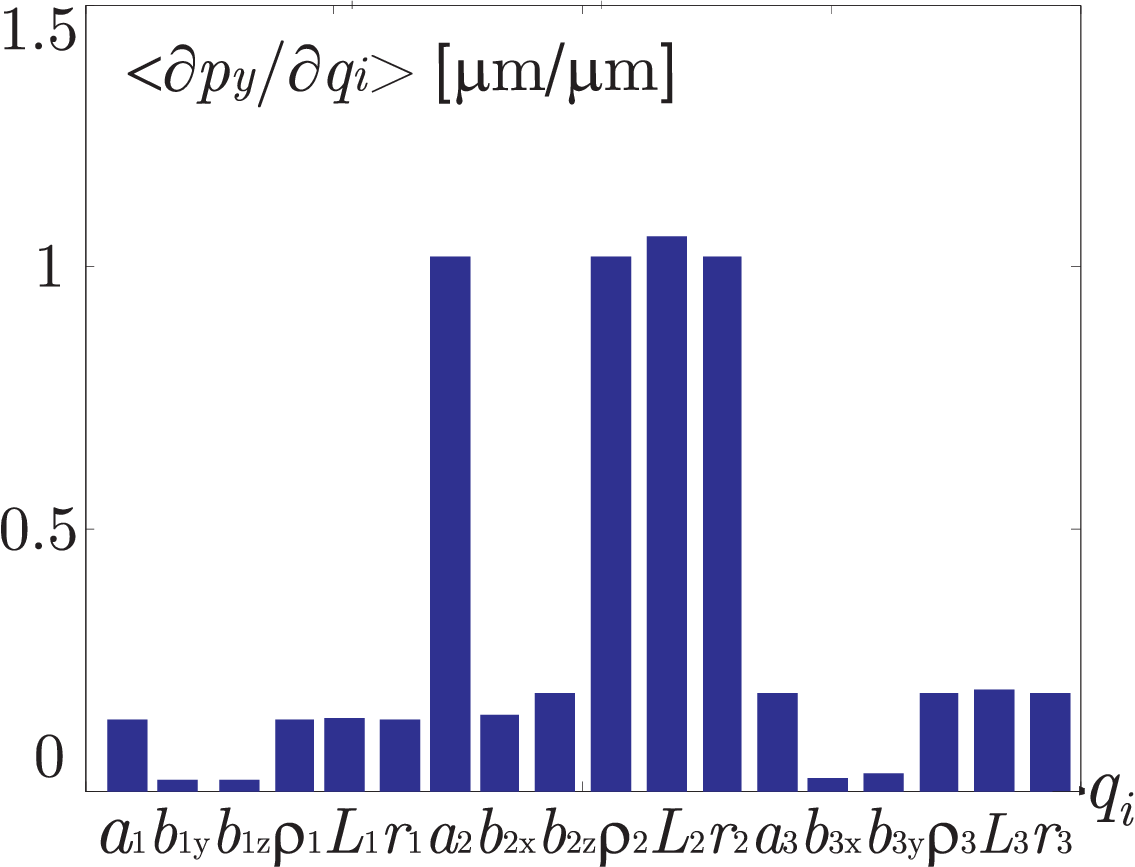}
\caption{Mean of sensitivity of $p_y$ throughout $C_u$}
\label{fig:sensi-py-Cu}
\end{figure}

\begin{figure}[!htbp]
\centering\includegraphics[width=60mm]{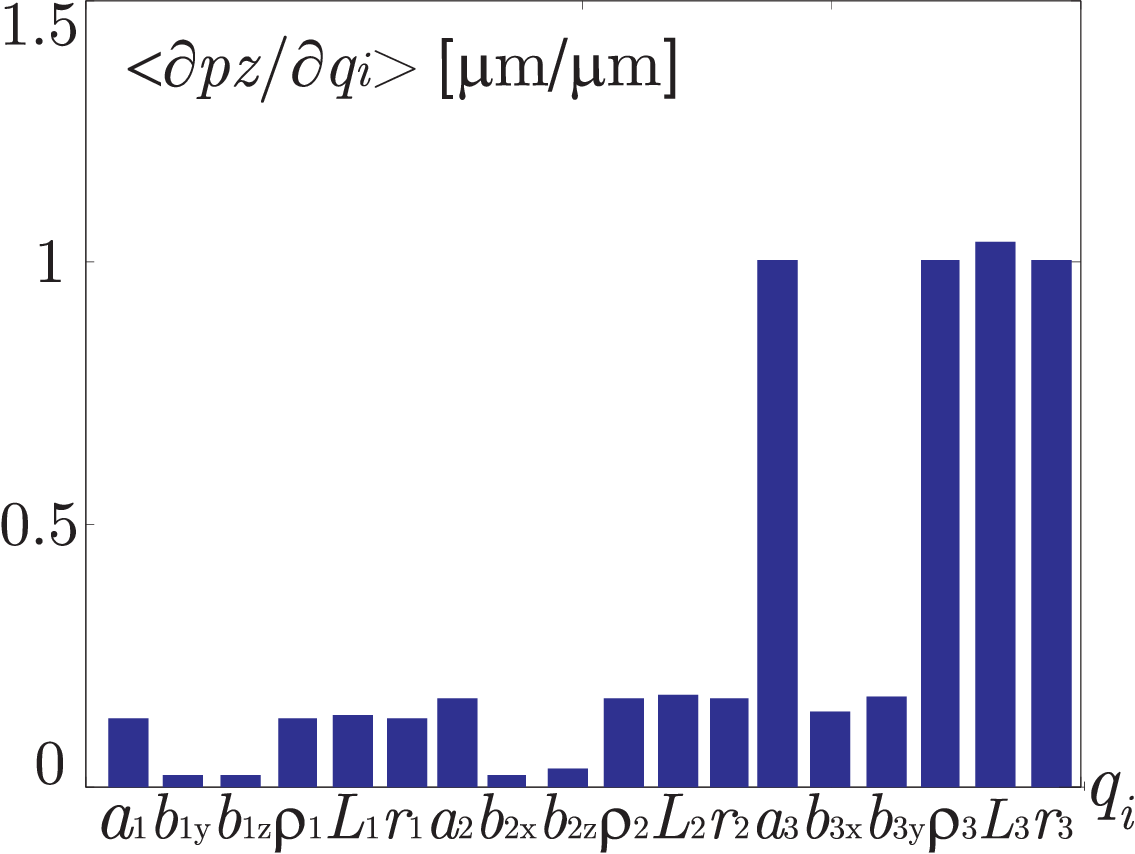}
\caption{Mean of sensitivity of $p_z$ throughout $C_u$}
\label{fig:sensi-pz-Cu}
\end{figure}

\begin{figure}[!htbp]
\centering\includegraphics[width=60mm]{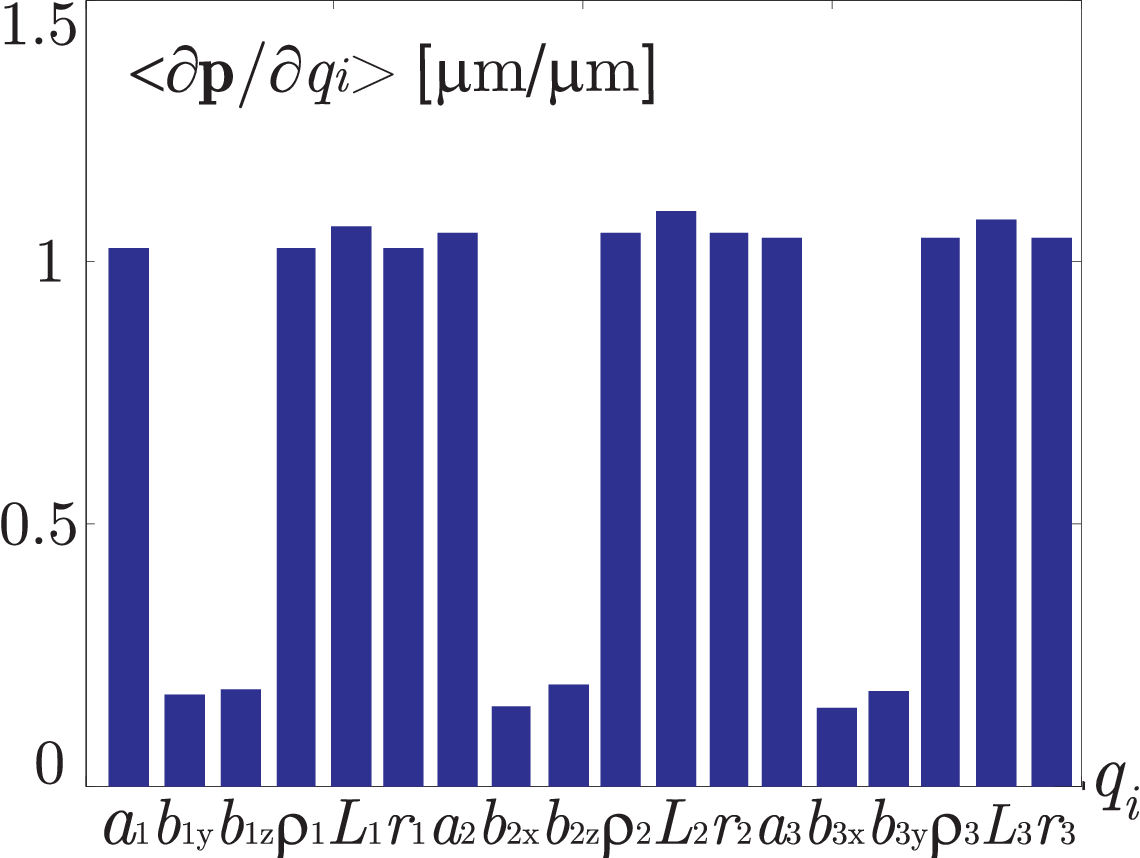}
\caption{Mean of sensitivity of {\bf p} throughout $C_u$}
\label{fig:sensi-p-Cu}
\end{figure}

%
%
%


Figures \ref{fig:sensi-px-Cu}, \ref{fig:sensi-py-Cu},
\ref{fig:sensi-pz-Cu} and \ref{fig:sensi-p-Cu} depict the mean of
the sensitivity coefficients of $p_x$, $p_y$, $p_z$, and ${\bf
p}$, when the end-effector covers $C_u$. It appears that the
position of the end-effector is very sensitive to variations in
the position of points $A_i$, variations in the lengths of the
parallelograms, $L_i$, variations in the lengths of prismatic
joints, $\rho_i$, and variations in the position of points $C_i$
defined by $r_i$ (see Fig.\ref{fig:schema-cinematique-jambe}).
However, it is little sensitive to the orientation errors of the
direction of the prismatic joints, defined by parameters
$b_{1y},b_{1z},b_{2x},b_{2z},b_{3x},b_{3y}$. Besides, it is
noteworthy that $p_x$ ($p_y$ , $p_z$, respectively) is very
sensitive to the design parameters which make up the $1^{st}$
($2^{nd}$, $3^{rd}$, respectively) leg of the manipulator,
contrary to the others. That is due to the symmetry of the
architecture of the manipulator. Henceforth, only the variations
in the design parameters of the first leg of the manipulator will
be taken into account. Indeed, the sensitivity of the position of
the end-effector to the variations in the design parameters of the
second and the third legs of the manipulator can be deduced from
the sensitivity of the position of the end-effector to variations
in the design parameters of the first leg.

Chablat et al. \cite{CHABLAT03} showed that if the prescribed
bounds of the velocity transmission factors (the kinematic
criteria used to dimension the manipulator) are satisfied at $Q_1$
and $Q_2$, then these bounds are satisfied throughout the
prescribed cubic Cartesian workspace $C_u$. $Q_1$ and $Q_2$ are
then the most critical points of $C_u$, whereas $O$ is the most
interesting point because it corresponds to the isotropic
kinematic configuration of the manipulator. Here, we assume that
if the prescribed bounds of the sensitivity coefficients are
satisfied at $Q_1$ and $Q_2$, then these bounds are satisfied
throughout $C_u$.

\begin{figure}[!htbp]
\centering\includegraphics[width=60mm]{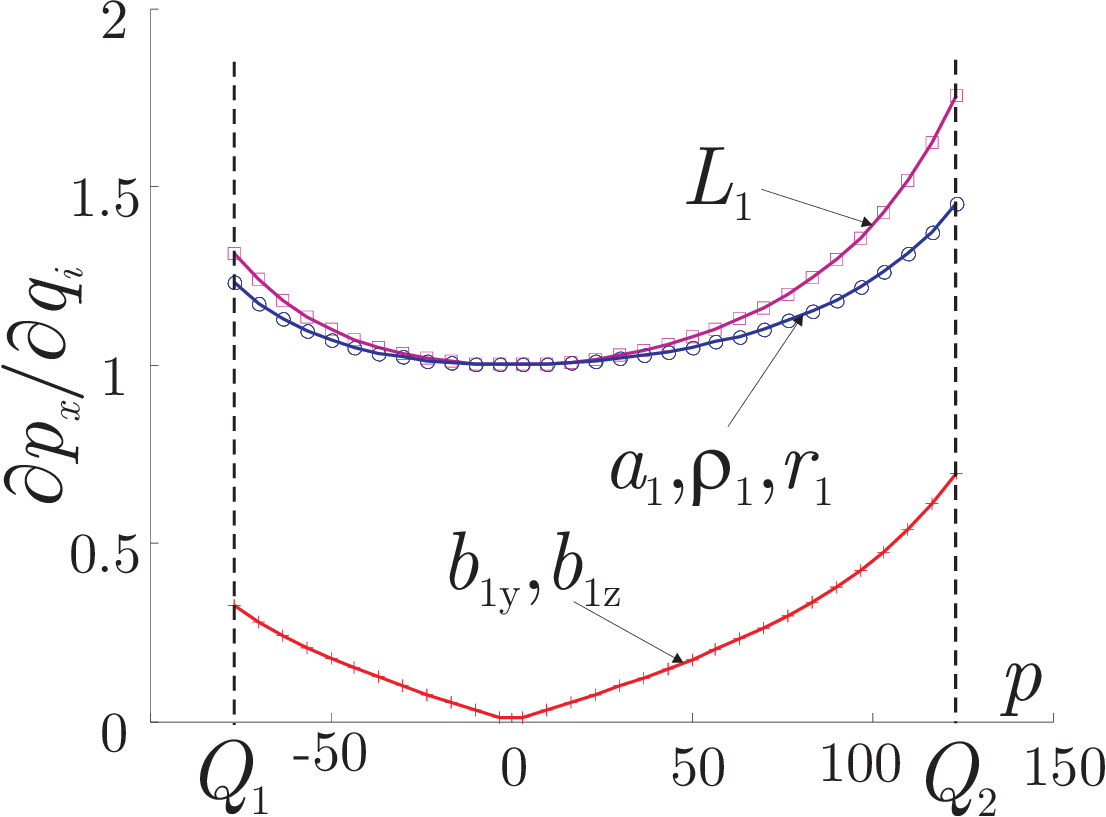}
\caption{Sensitivity of $p_x$ to the variations in the $1^{st}$
leg} \label{fig:influence-vardim-px}
\end{figure}

\begin{figure}[!htbp]
\centering\includegraphics[width=60mm]{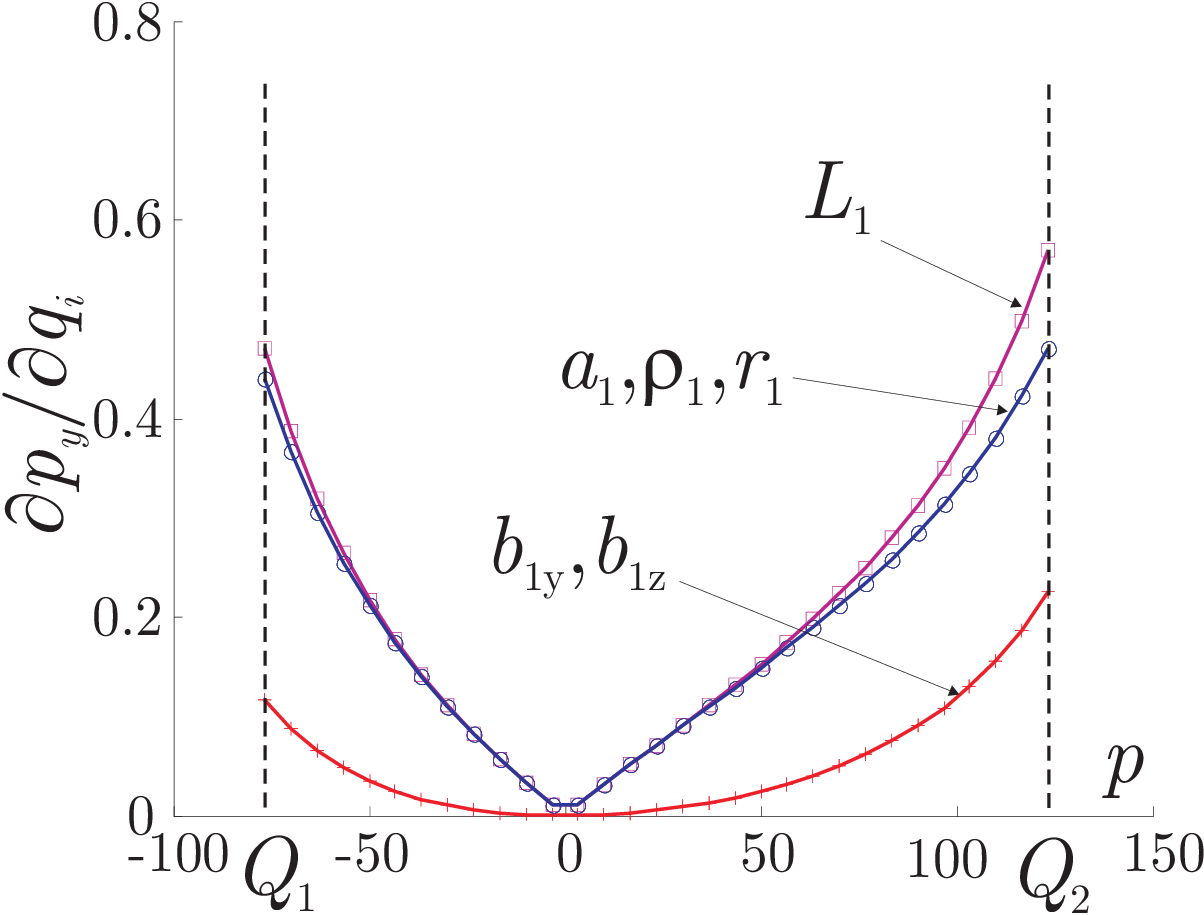}
\caption{Sensitivity of $p_y$ to the variations in the $1^{st}$
leg} \label{fig:influence-vardim-py}
\end{figure}

\begin{figure}[!htbp]
\centering\includegraphics[width=60mm]{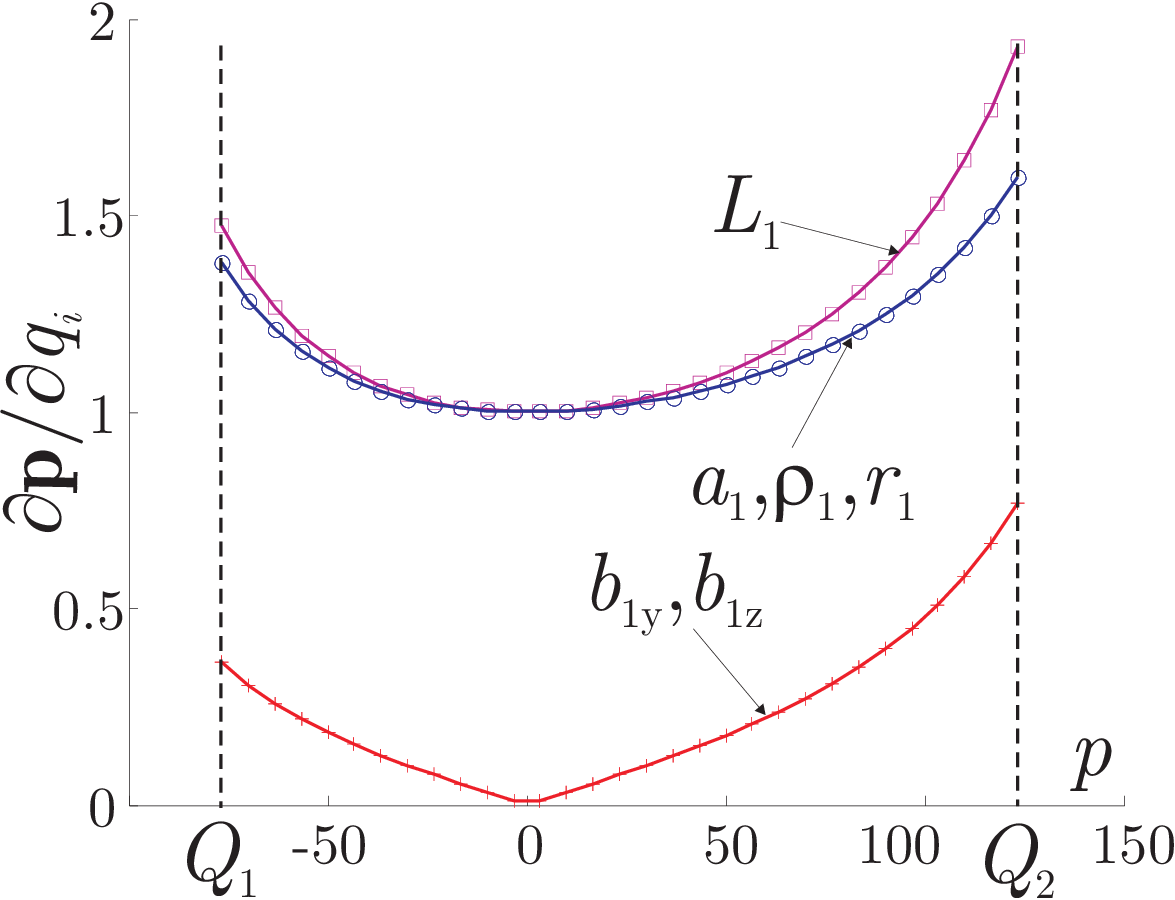}
\caption{Sensitivity of ${\bf p}$ to the variations in the
$1^{st}$ leg} \label{fig:influence-vardim-p}
\end{figure}

\begin{figure}[!htbp]
\centering\includegraphics[width=60mm]{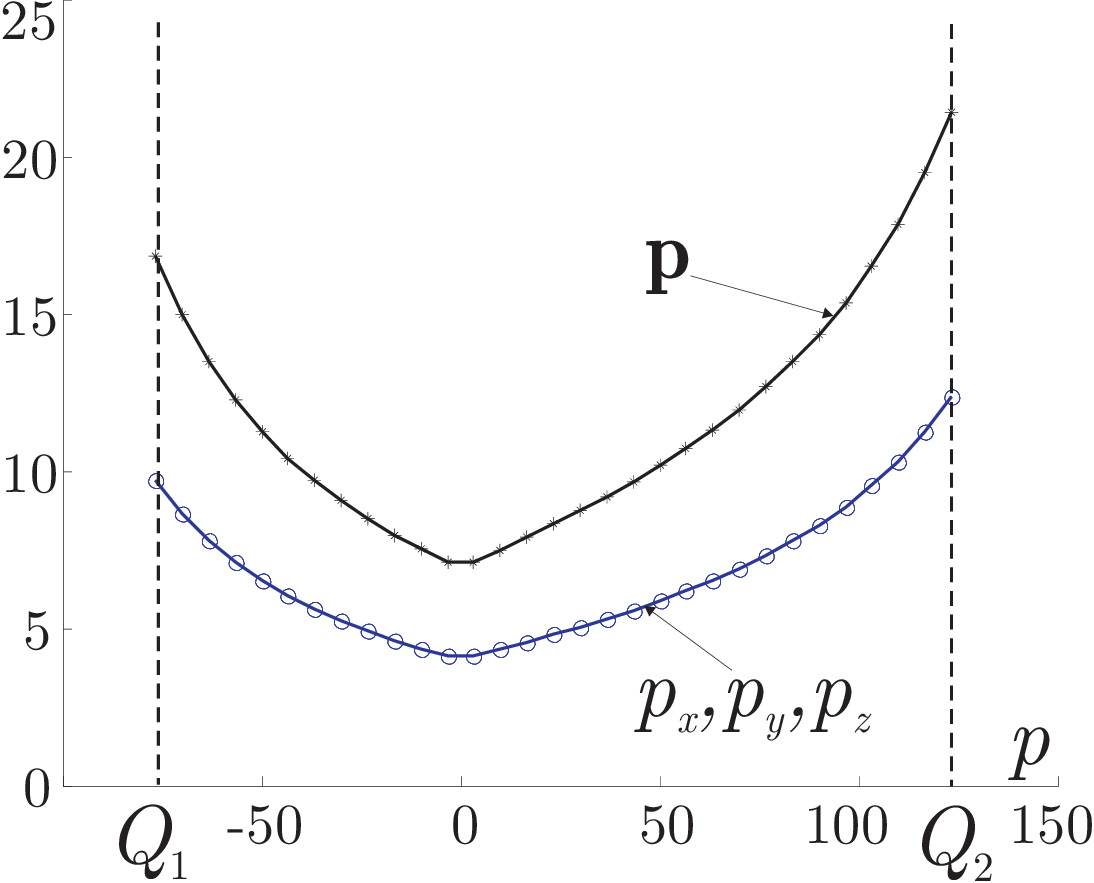}
\caption{Global sensitivity of ${\bf p}$, $p_x$, $p_y$, and $p_z$}
\label{fig:influence_pxpypzp}
\end{figure}

%
%
%

Figures \ref{fig:influence-vardim-px} and
\ref{fig:influence-vardim-py} depict the sensitivity coefficients
of $p_x$ and $p_y$ to the dimensional variations in the $1^{st}$
leg, {\it i.e.}: $a_1,b_{1y},b_{1z},\rho_1,L_1,r_1$, along
$Q_1Q_2$. It appears that these coefficients are a minimum in the
isotropic configuration, {\it i.e.}: $P \equiv O$, and a maximum
when $P \equiv Q_2$, {\it i.e.}: in the closest configuration to
the singular one. Figure \ref{fig:influence-vardim-p} depicts the
sensitivity coefficients of ${\bf p}$ along diagonal $Q_1Q_2$. It
is noteworthy that all the sensitivity coefficients are a minimum
when $P \equiv O$ and a maximum when $P \equiv Q_2$. Finally,
figure \ref{fig:influence_pxpypzp} depicts the global
sensitivities of $\bf p$, $p_x$, $p_y$, and $p_z$ to the
dimensional variations. It appears that they are a minimum when $P
\equiv O$, and a maximum when $P \equiv Q_2$.

\begin{figure}[!htbp]
\centering\includegraphics[width=60mm]{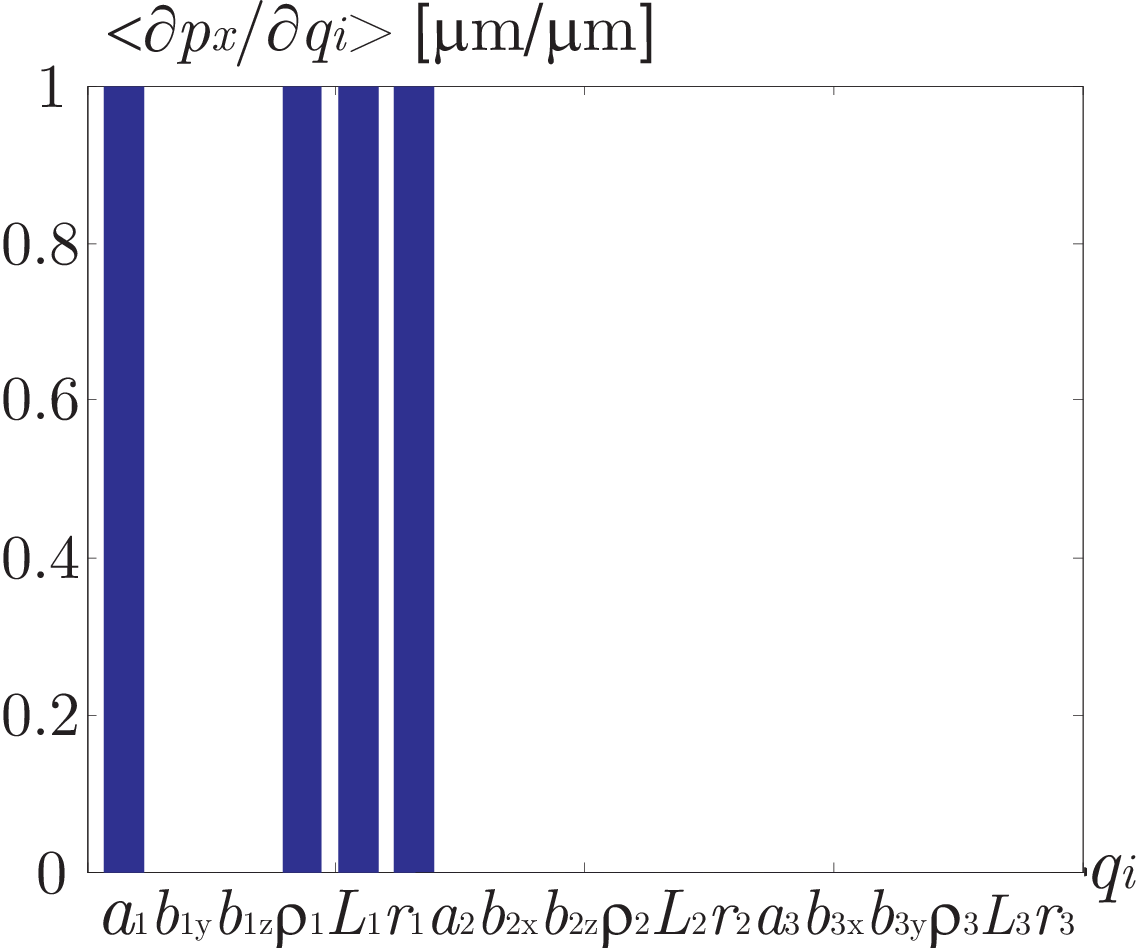}
\caption{Sensitivity of $p_x$ in the isotropic configuration}
\label{fig:rob-iso-sensi-px}
\end{figure}

\begin{figure}[!htbp]
\centering\includegraphics[width=60mm]{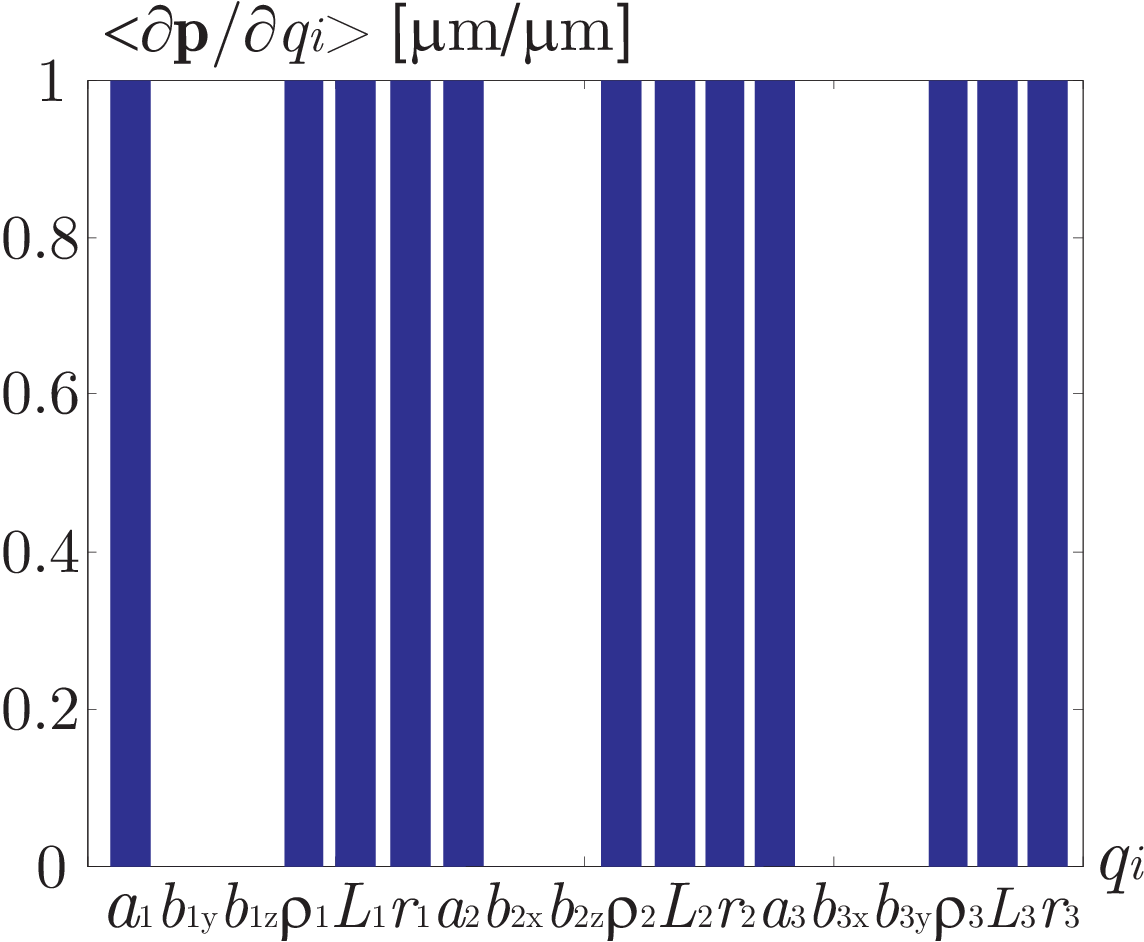}
\caption{Sensitivity of $\bf p$ in the isotropic configuration}
\label{fig:rob-iso-sensi-p}
\end{figure}

%
%

Figures \ref{fig:rob-iso-sensi-px} and \ref{fig:rob-iso-sensi-p}
depict the sensitivity coefficients of $p_x$ and $\bf p$ in the
isotropic configuration. In this configuration, the position error
of the end-effector does not depend on the orientation errors of
the directions of the prismatic joints because the sensitivity of
the position of $P$ to variations in
$b_{1y},b_{1z},b_{2x},b_{2z},b_{3x},b_{3y}$ is null in this
configuration. Besides, variations in $p_x$, $p_y$, and $p_z$ are
decoupled in this configuration. Indeed, variarions in $p_x$,
($p_y$, $p_z$, respectively) are only due to dimensional
variations in the $1^{st}$, ($2^{nd}$, $3^{rd}$, respectively) leg
of the manipulator. The corresponding sensitivity coefficients are
equal to 1. It means that the dimensional variations are neither
amplified nor compensated in the isotropic configuration.

\begin{figure}[!htbp]
\centering\includegraphics[width=60mm]{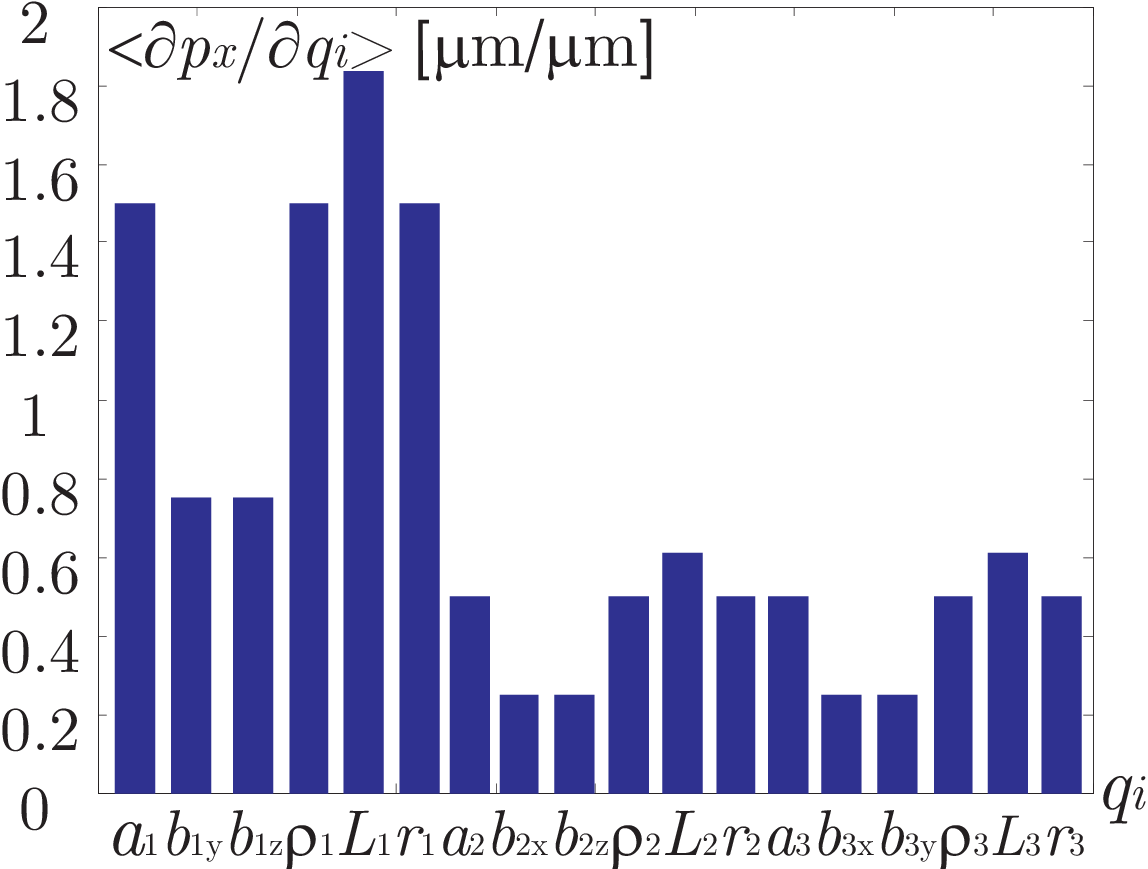}
\caption{$Q_2$ configuration, sensitivity of $p_x$}
\label{fig:rob-Q2-sensi-px}
\end{figure}

\begin{figure}[!htbp]
\centering\includegraphics[width=60mm]{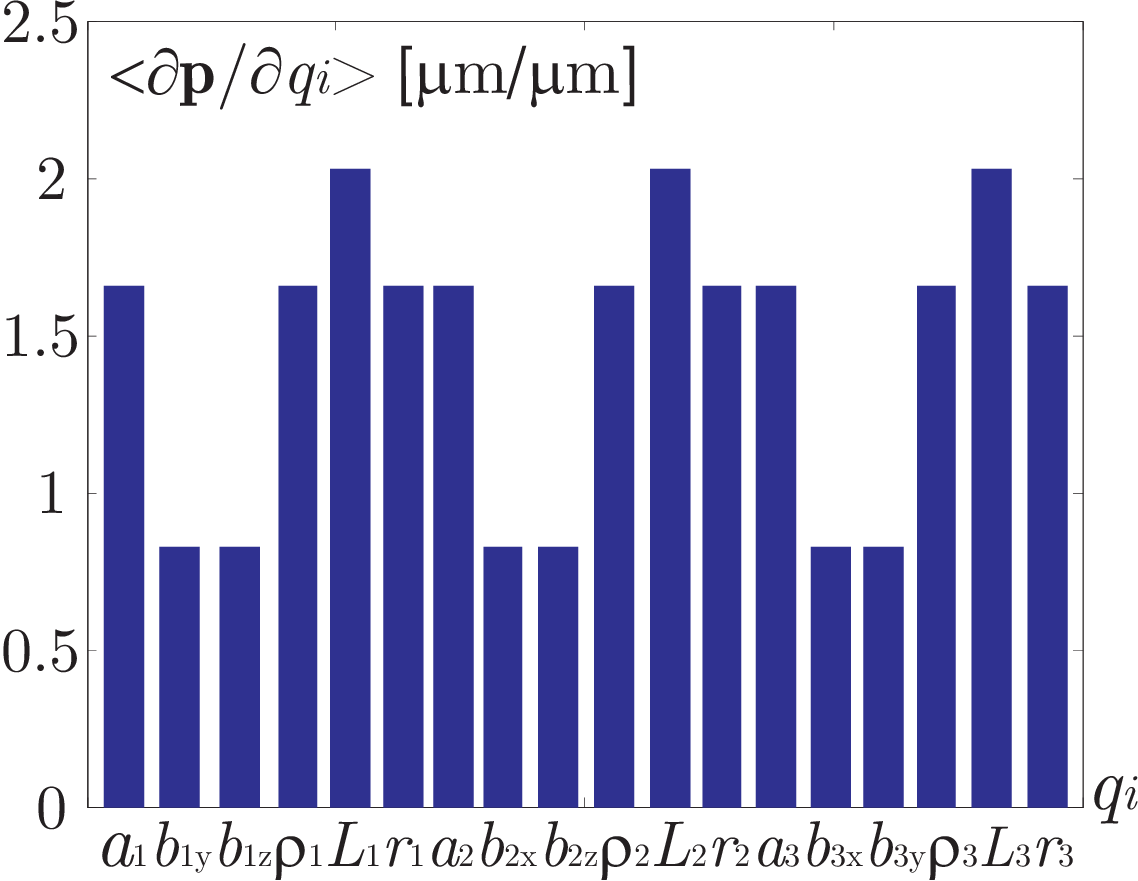}
\caption{$Q_2$ configuration, sensitivity of ${\bf p}$}
\label{fig:rob-Q2-sensi-p}
\end{figure}

%
%

Figures \ref{fig:rob-Q2-sensi-px} and \ref{fig:rob-Q2-sensi-p}
depict the sensitivity coefficients of $p_x$ and $\bf p$ when the
end-effector hits $Q_2$ ($P \equiv Q_2$). In this case, variations
in $p_x$, $p_y$, and $p_z$ are coupled. For example, variations in
$p_x$ are due to both dimensional variations in the $1^{st}$ leg
and variations in the $2^{nd}$ and the $3^{rd}$ legs. Besides, the
amplification of the dimensional variations is important. Indeed,
the sensitivity coefficients of $\bf p$ are close to 2 in this
configuration. For example, as the sensitivity coefficient
relating to $L_1$ is equal to 1.9, the position error of the
end-effector will be equal to $19 \mu m$ if $\delta L_1 = 10\mu
m$. Moreover, we noticed numerically that $Q_2$ configuration is
the most sensitive configuration to dimensional variations of the
manipulator.

According to figures \ref{fig:sensi-px-Cu} - \ref{fig:sensi-p-Cu},
\ref{fig:rob-iso-sensi-px} - \ref{fig:rob-Q2-sensi-p}, variations
in design parameters of the same type from one leg to another have
the same influence on the location of the end-effector.

However, this linkage kinematic method does not take into account
variations in the parallelograms, except the variations in their
global length. Thus, a differential vector method is developed
below.

\subsection{Differential Vector Method}
\label{section:differential-vector-method}
In this section, we perfect a sensitivity analysis method of the
Orthoglide, which complements the previous one. This method is
used to analyze the sensitivity of the position and the
orientation of the end-effector to dimensional and angular
variations, and particularly to the variations in the
parallelograms. Moreover, it allows us to distinguish the
variations which are responsible for the position errors of the
end-effector from the ones which are responsible for its
orientation errors. To develop this method, we were inspired by a
Huang \& al. work on a parallel kinematic machine, which is made
up of parallelogram joints too \cite{HUANG03}.

First, we express the dimensional and angular variations in
vectorial form. Then, a relation between the position and the
orientation errors of the end-effector is obtained from the
closed-loop kinematic equations. The expressions of the
orientation and the position errors of the end-effector, with
respect to the variations in the design parameters, are deduced
from this relation. Finally, we introduce two sensitivity indices
to assess the sensitivity of the position and the orientation of
the end-effector to dimensional and angular variations, and
particularly to the parallelism errors of the bars of the
parallelograms.

\begin{figure}[!h]
\begin{center}
\includegraphics[width=80mm]{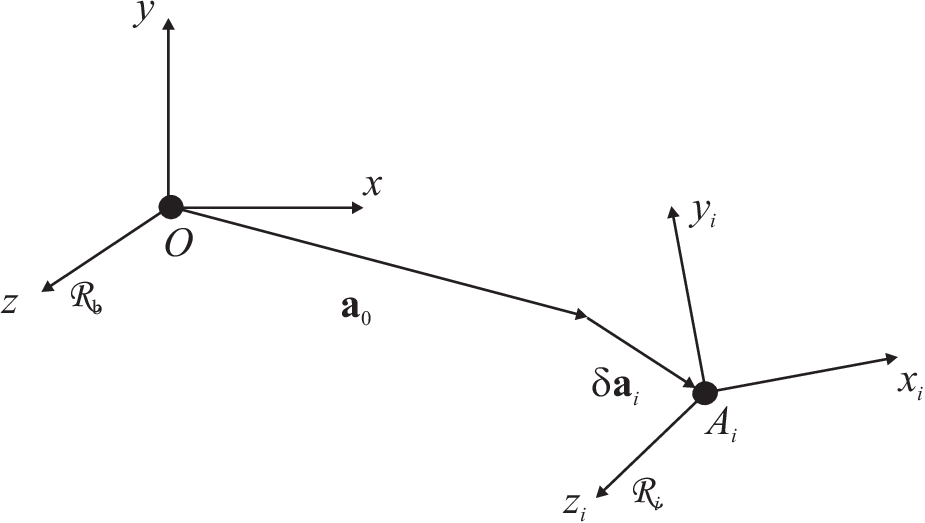}
\caption{Variations in $O-A_i$ chain} \label{fig:OAi}
\end{center}
\end{figure}

\subsubsection{Formulation}
\label{section:differential-vector-method-theory}
\begin{figure}[!h]
\begin{center}
\includegraphics[width=80mm]{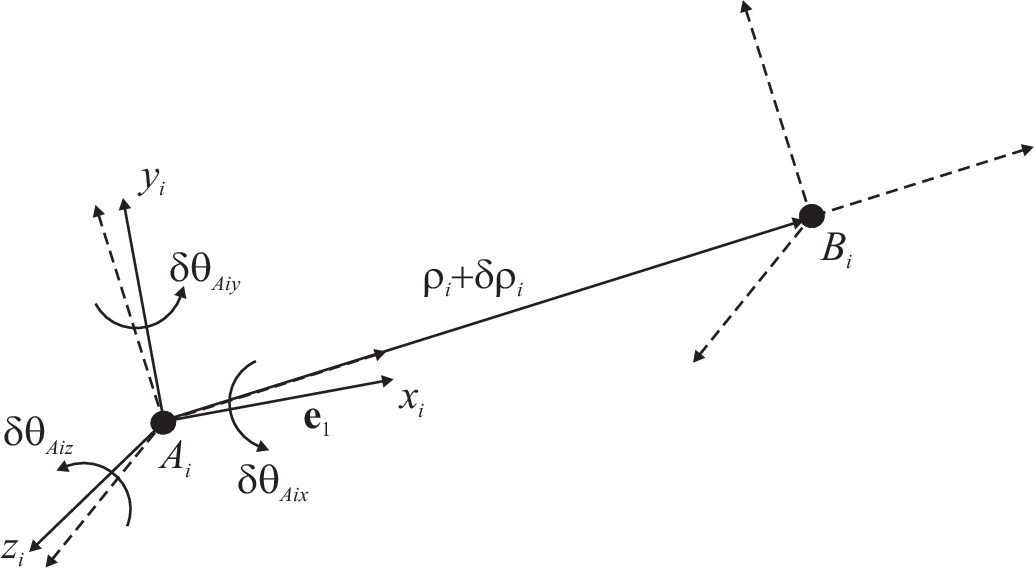}
\caption{Variations in $A_i-B_i$ chain} \label{fig:AiBi}
\end{center}
\end{figure}

\begin{figure}[!h]
\begin{center}
\includegraphics[width=80mm]{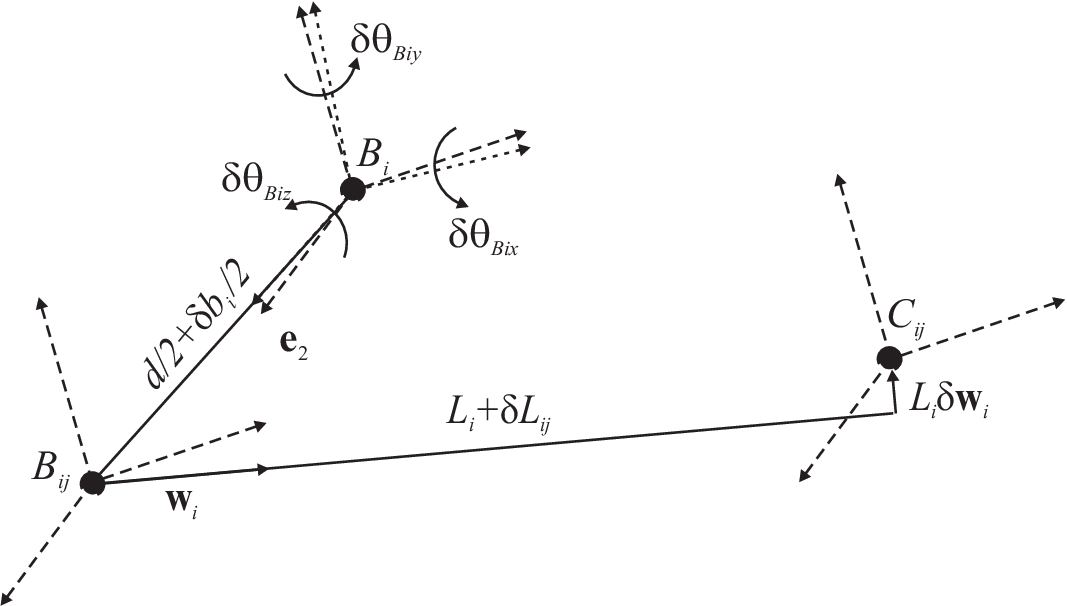}
\caption{Variations in $B_i-B_{ij}-C_{ij}$ chain}
\label{fig:BiBijCij}
\end{center}
\end{figure}

\begin{figure}[!h]
\begin{center}
\includegraphics[width=80mm]{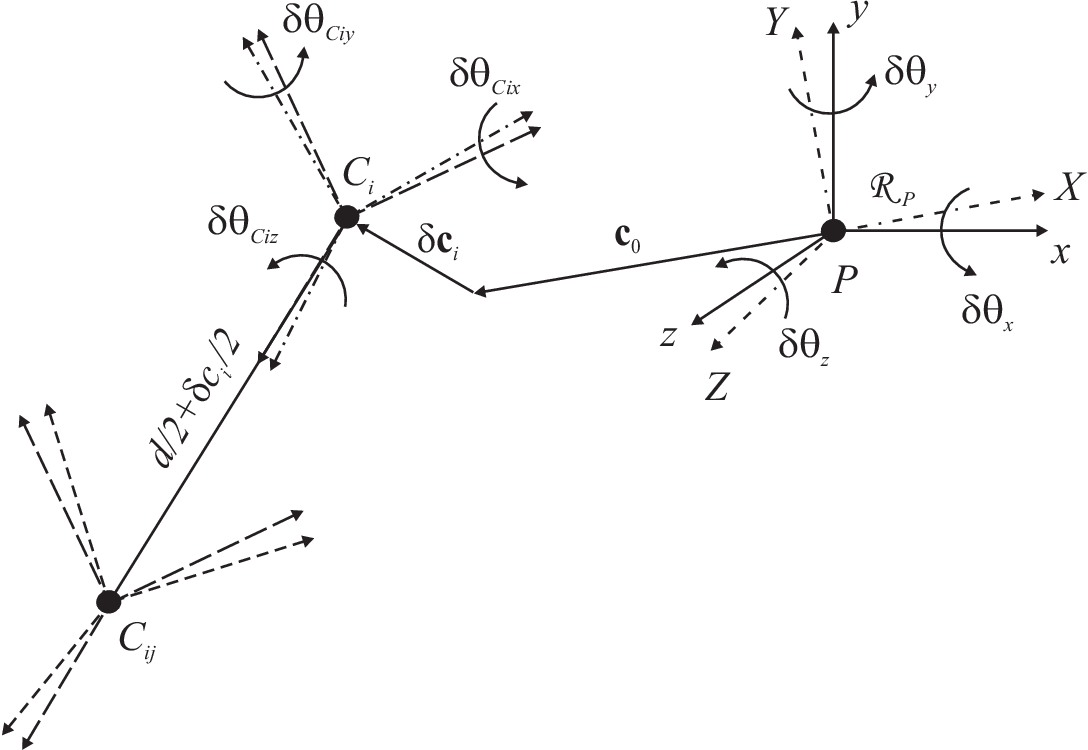}
\caption{Variations in $C_{ij}-C_i-P$ chain} \label{fig:CijCiP}
\end{center}
\end{figure}

The schematic drawing of the $i^{th}$ leg of the Orthoglide
depicted in Fig.\ref{fig:schema-cinematique-jambe} is split in
order to depict the variations in design parameters in a vectoriel
form. The closed-loop kinematic chains
$O-A_i-B_i-B_{ij}-C_{ij}-C_{i}-P$, $i=1,2,3$, $j=1,2$, are
depicted by Figs.\ref{fig:OAi}-\ref{fig:CijCiP}. ${\cal R}_i$ is
the coordinate frame attached to the $i^{th}$ prismatic joint.
${\bf o}, {\bf a}_i, {\bf b}_i, {\bf b}_{ij}, {\bf c}_{ij}, {\bf
c}_{i}, {\bf p}$, are the Cartesian coordinates of points $O, A_i,
B_i, B_{ij}, C_{ij}, C_{i}, P$, respectively, expressed in ${\cal
R}_i$ and depicted in Fig.\ref{fig:schema-cinematique-jambe}.

According to Fig.\ref{fig:OAi},
\begin{equation}\label{eq:OAi}
  {\bf a}_i - {\bf o} = {\bf R}_i ({\bf a}_0 + \delta{\bf a}_i)
\end{equation}

where ${\bf a}_0$ is the nominal position vector of
$A_i$ with respect to $O$ expressed in ${\cal R}_i$, $\delta{\bf
a}_i$ is the positioning error of $A_i$. ${\bf R}_i$ is the
transformation matrix from ${\cal R}_i$ to ${\cal R}_b$. ${\bf
I}_3$ is the ($3\times3$) identity matrix and
\begin{eqnarray}\label{eq:R1R2R3}
    {\bf R}_1 & = & {\bf I}_3 \\
    {\bf R}_2 & = & \mat{ccc}{0 & 0 & -1 \\ 1 & 0 & 0 \\ 0 & -1 &  0} \\
    {\bf R}_3 & = & \mat{ccc}{0 & 1 & 0 \\ 0 & 0 & 1 \\ 1 & 0 & 0}
\end{eqnarray}

According to Fig.\ref{fig:AiBi},
\begin{equation}\label{eq:AiBi} {\bf b}_i - {\bf a}_i = {\bf R}_i
( \rho_i + \delta \rho_i ) {\bf e}_1 + {\bf R}_i {\delta
\theta}_{Ai} \times ( \rho_i + \delta \rho_i ) {\bf e}_1
\end{equation}

where $\rho_i$ is the displacement of the $i^{th}$
prismatic joint, $\delta \rho_i$ is its displacement error,
$\delta \theta_{Ai}= \mat{ccc}{\delta \theta_{Aix} & \delta
\theta_{Aiy} & \delta \theta_{Aiz}}^{T}$ is the angular variation
of its direction, and
\begin{eqnarray}\label{eq:R1R2R3}
    {\bf e}_1 & = & \mat{c}{1 \\ 0 \\ 0} \\
    {\bf e}_2 & = & \mat{c}{0 \\ 0 \\ 1} \\
    \xi(j) & = & \left\{\begin{array}{cc} 1 & \textrm{if } j=1 \\ -1 & \textrm{if } j=2
    \end{array}\right.
\end{eqnarray}

According to Fig.\ref{fig:BiBijCij},
\begin{eqnarray}\label{eq:BiBijCij}
{\bf b}_{ij} - {\bf b}_i & = & {\bf R}_i [{\bf I}_3 + {\delta
\theta}_{Ai} \times] \big( \xi(j) (d/2 + \delta {b}_i /2) \nonumber \\
& & [{\bf I}_3 + \delta {\theta}_{Bi} \times] {\bf e}_2 \big)\\
{\bf c}_{ij} - {\bf b}_{ij} & = & L_i {\bf w}_i + \delta L_{ij}
{\bf w}_i + L_i \delta {\bf w}_i
\end{eqnarray}
\noindent where $d$ is the nominal width of the parallelogram,
$\delta {b}_i$ is the variation in the length of link
$\overline{B_{i1}B_{i2}}$ and is supposed to be equally shared by
each side of $B_i$. $\delta \theta_{Bi}= \mat{ccc}{\delta
\theta_{Bix} & \delta \theta_{Biy} & \delta \theta_{Biz}}^{T}$ is
the orientation error of link $\overline{B_{i1}B_{i2}}$ with
respect to the direction of the $i^{th}$ prismatic joint, $L_i$ is
the length of the $i^{th}$ parallelogram, $\delta L_{ij}$ is the
variation in the length of link $\overline{B_{ij}C_{ij}}$, of
which ${\bf w}_i$ is the direction, and $\delta {\bf w}_i$ is the
variation in this direction, orthogonal to ${\bf w}_i$.

According to Fig.\ref{fig:CijCiP},
\begin{eqnarray} \label{eq:PCij} {\bf c}_{ij} - {\bf c}_{i} & = & {\bf R}_i [{\bf I}_3 + \delta \theta \times] \big(
\xi(j)(d/2 + \delta{c}_i /2) \nonumber \\ & & [{\bf I}_3 + \delta
{\theta}_{Ci} \times]{\bf e}_2 \big)\\ {\bf c}_{i} - {\bf p} & = &
[{\bf I}_3 + \delta \theta \times] {\bf R}_i ({\bf c}_0 +
\delta{\bf c}_i)
\end{eqnarray}

\noindent where $\delta {c}_i$ is the variation in the length of
link $\overline{C_{i1}C_{i2}}$, which is supposed to be equally
shared by each side of $C_i$. $\delta \theta_{Ci} =
\mat{ccc}{\delta \theta_{Cix} & \delta \theta_{Ciy} & \delta
\theta_{Ciz}}^{T}$ is the orientation error of link
$\overline{C_{i1}C_{i2}}$ with respect to link $\overline{C_iP}$.
${\bf c}_0$ is the nominal position vector of $C_i$ with respect
to end-effector $P$, expressed in ${\cal R}_i$, $\delta{\bf c}_i$
is the position error of $C_i$ expressed in ${\cal R}_i$, and
$\delta \theta = \mat{ccc}{\delta \theta_x & \delta \theta_y &
\delta \theta_z}^{T}$ is the orientation error of the
end-effector, expressed in ${\cal R}_b$.

\begin{figure}[!htbp]
\begin{center}
\centering\includegraphics[width=80mm]{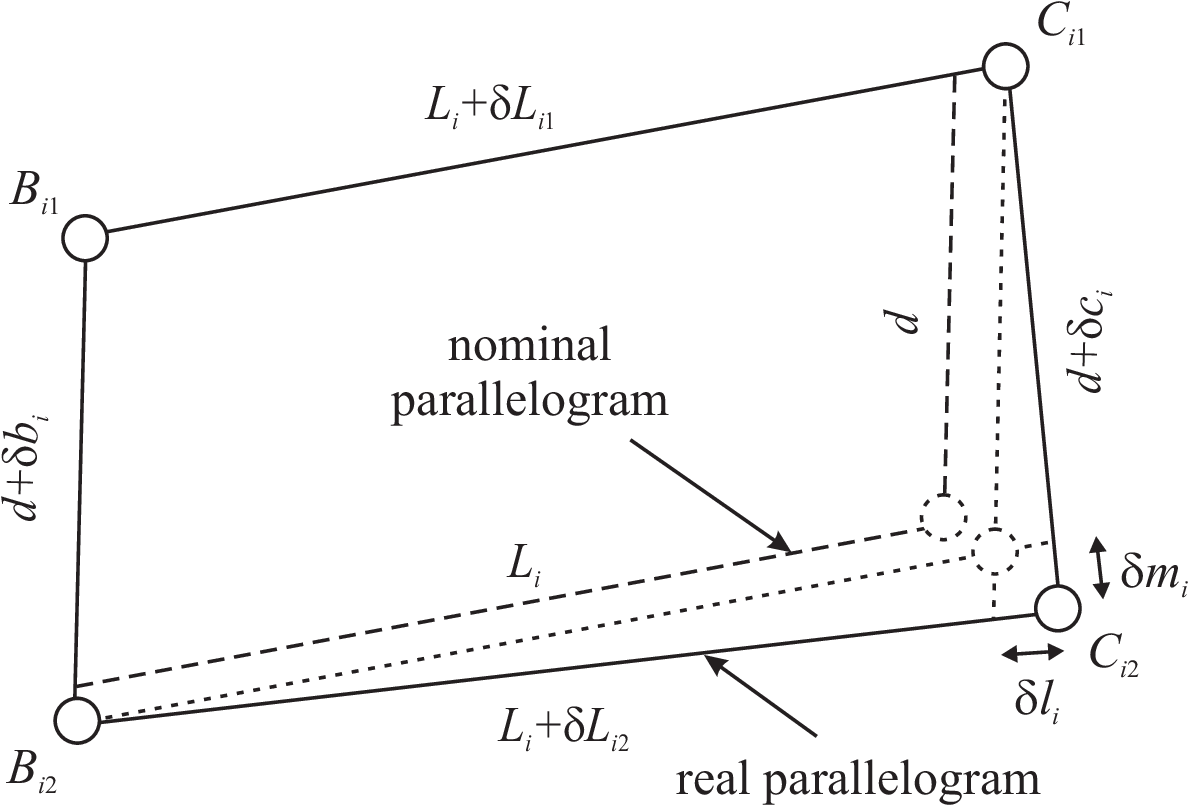}
\caption{Variations in the $i^{th}$ parallelogram}
\label{fig:parallelogramme}
\end{center}
\end{figure}

Implementing linearization of eqs.(\ref{eq:OAi}-\ref{eq:PCij}) and
removing the components associated with the nominal constrained
equation ${\bf p}_0 = {\bf R}_i ({\bf a}_0 + \rho_i {\bf e}_1 -
{\bf c}_0) + L_i {\bf w}_i$, yields
\begin{eqnarray}\label{eq:delta-r} \delta {\bf p} & = & {\bf p} - {\bf p}_0 \nonumber\\ & = & {\bf R}_i
\big( \delta {\bf e}_i + \rho_i (\delta \theta_{Ai} \times {\bf
e}_1) + \xi(j) \ d/2 \ ( \delta \theta_{Ai} \times {\bf e}_2) + \\
& & \xi(j) \ d/2 \
(\delta \gamma_i \times {\bf e}_2) + \xi(j) \ \delta{m_i}/2 \ {\bf e}_2 \big) + \nonumber\\
& & \delta L_{ij} {\bf w}_i + L_{i} \delta {\bf w}_i - \delta
\theta \times {\bf R}_i \big( {\bf c}_0 + d/2 \ \xi(j) \ {\bf e}_2
\big) \nonumber
\end{eqnarray}

\noindent where
\begin{description}
\item $\delta {\bf p}$ is the position error of the end-effector
of the manipulator.
\item $\delta {\bf e}_i = \delta {\bf a}_i +
\delta{\rho}_i {\bf e}_1 - \delta {\bf c}_i$ is the sum of the
position errors of points $A_i$, $B_i$, and $C_i$ expressed in
${\cal R}_i$.
\item $\delta \gamma_i = \delta \theta_{Bi} - \delta
\theta_{Ci}$ is the sum of the orientation errors of the $i^{th}$
parallelogram with respect to the $i^{th}$ prismatic joint and the
end-effector.
\item $\delta{m_i} = \delta{b}_i - \delta{c}_i $
corresponds to the parallelism error of links
$\overline{B_{i1}C_{i1}}$ and $\overline{B_{i2}C_{i2}}$, which is
depicted by Fig.\ref{fig:parallelogramme}.
\end{description}

Equation (\ref{eq:delta-r}) shows the coupling of the position and
orientation errors of the end-effector. Contrary to the
orientation error, the position error can be compensated because
the manipulator is a translational 3-DOF PKM. Thus, it is more
important to minimize the geometrical variations, which are
responsible for the orientation errors of the end-effector than
the ones, which are responsible for its position errors.

The following equation is obtained by multiplying both sides of
eq.(\ref{eq:delta-r}) by ${\bf w}_i^T$ and utilizing the
circularity of hybrid product.
\begin{eqnarray}\label{eq:delta-r-2}
{\bf w}_i^T \delta {\bf p} & = & {\bf w}_i^T {\bf R}_i {\delta
{\bf e}_i} + \rho_i ({\bf R}_i {\bf e}_1 \times {\bf w}_i)^T {\bf
R}_i \delta \theta_{Ai} + \xi(j) \ d/2  \\ &  & ({\bf R}_i {\bf
e}_2 \times {\bf w}_i)^T {\bf R}_i (\delta \theta_{Ai} + \delta
\gamma_i) + \xi(j) \ \delta{m_i}/2 \ {\bf w}_i^T {\bf R}_i {\bf
e}_2 \nonumber \\ & & + \delta L_{ij} - \big( {\bf R}_i ({\bf c}_0
+ \xi(j) \ d/2 \ {\bf e}_2) \times {\bf w}_i \big)^T \delta \theta
\nonumber
\end{eqnarray}

{\bf Orientation Error Mapping Function:} By substraction of
eqs.(\ref{eq:delta-r-2}) written for $j=1$ and $j=2$, and for the
$i^{th}$ kinematic chain, a relation is obtained between the
orientation error of the end-effector and the variations in design
parameters, which is independent of the position error of the
end-effector.
\begin{equation}\label{eq:orientation-error-mapping-function}
d({\bf R}_i {\bf e}_2 \times {\bf w}_i)^T \delta \theta = \delta
l_i + d({\bf R}_i {\bf e}_2 \times {\bf w}_i)^T {\bf R}_i (\delta
\theta_{Ai} + \delta \gamma_i) + \delta{m_i} \ {\bf w}_i^T {\bf
R}_i {\bf e}_2
\end{equation}

\noindent where $\delta l_i = \delta L_{i1} - \delta L_{i2}$, the
relative length error of links $\overline{B_{i1}C_{i1}}$ and
$\overline{B_{i2}C_{i2}}$, depicts the parallelism error of links
$\overline{B_{i1}B_{i2}}$ and $\overline{C_{i1}C_{i2}}$ as shown
in Fig.\ref{fig:parallelogramme}. Equation
(\ref{eq:orientation-error-mapping-function}) can be written in
matrix form:
\begin{equation}\label{eq:expression-Jtheta-theta}
  \delta \theta = {\bf J}_{\theta \theta} {\gbf \epsilon}_{\theta}
\end{equation}

\noindent with
\begin{eqnarray}
  {\bf J}_{\theta \theta} &=& {\bf D}^{-1}{\bf E} \\
  {\bf D} &=& d \mat{c}{({\bf R}_1 {\bf e}_2 \times {\bf w}_1)^T \\ ({\bf R}_2 {\bf e}_2 \times {\bf w}_2)^T \\ ({\bf R}_3 {\bf e}_2 \times {\bf
  w}_3)^T} \\ {\bf E} &=& \mat{ccc}{{\bf E}_1 & \cdots & \cdots \\ \cdots  & {\bf E}_2 & \cdots \\ \cdots & \cdots & {\bf
  E}_3}\\
  {\bf E}_i &=& \mat{cccc}{1 & {\bf w}_i^T {\bf R}_i{\bf e}_2 & d({\bf R}_i{\bf e}_2 \times {\bf w}_i)^T{\bf R}_i & d({\bf R}_i{\bf e}_2 \times {\bf w}_i)^T{\bf R}_i}
\end{eqnarray}
$\delta \theta$ is the orientation error of the end-effector
expressed in ${\cal R}_b$, and ${\gbf \epsilon}_{\theta} = ({\gbf
\epsilon}_{\theta1}^T, {\gbf \epsilon}_{\theta2}^T, {\gbf
\epsilon}_{\theta3}^T )^T$ such that ${\gbf \epsilon}_{\theta i} =
(\delta l_i, \delta m_i, \delta \theta_{Ai}^T, \delta
\gamma_i^T)^T$. The determinant of ${\bf D}$ will be null if the
normal vectors to the plans, which contain the three
parallelograms respectively, are collinear, or if one
parallelogram is flat. Here, this determinant is not null when $P$
covers $C_u$ because of the geometry of the manipulator.
Therefore, ${\bf D}$ is nonsingular and its inverse ${\bf D}^{-1}$
exists.

As $({\bf R}_i {\bf e}_2 \times {\bf w}_i)^T \bot {\bf R}_i {\bf
e}_2$, $\delta \theta_{Aiz}$ and $\delta \gamma_{iz}$, the third
components of $\delta \theta_{Ai}$ and $\delta \gamma_{i}$
expressed in ${\cal R}_i$, have no effect on the orientation of
the end-effector. Thus, matrix ${\bf J}_{\theta \theta}$ can be
simplified by eliminating its columns associated with $\delta
\theta_{Aiz}$ and $\delta \gamma_{iz}$, $i=1,2,3$. Finally,
eighteen variations: $\delta l_i$, $\delta m_i$, $\delta
\theta_{Aix}$, $\delta \theta_{Aiy}$, $\delta \gamma_{ix}$,
$\delta \gamma_{iy}$, $i=1,2,3$, should be responsible for the
orientation error of the end-effector.\\

{\bf Position Error Mapping Function:} By addition of
eqs.(\ref{eq:delta-r-2}) written for $j=1$ and $j=2$, and for the
$i^{th}$ kinematic chain, a relation is obtained between the
position error of the end-effector and the variations in design
parameters, which does not depend on $\delta \gamma_{i}$.
\begin{equation}\label{eq:position-error-mapping-function}
{\bf w}_i^T \delta{\bf p} = \delta L_i + {\bf w}_i^T {\bf R}_i
\delta {\bf e}_i + \rho_i({\bf R}_i {\bf e}_1 \times {\bf w}_i)^T
{\bf R}_i \delta \theta_{Ai} - ({\bf R}_i {\bf c}_0 \times {\bf
w}_i)^T \delta \theta
\end{equation}

\noindent Equation (\ref{eq:position-error-mapping-function}) can
be written in matrix form:
\begin{equation}\label{delta-r}
  \delta{\bf p} = {\bf J}_{pp} {\gbf \epsilon}_{p} + {\bf J}_{p \theta} {\gbf
  \epsilon}_{\theta} = [{\bf J}_{pp} {\bf J}_{p \theta}]\left[\begin{array}{c} {\gbf \epsilon}_{p} \\ {\gbf
  \epsilon}_{\theta} \end{array} \right]
\end{equation}

\noindent with
\begin{eqnarray}
{\bf J}_{pp} & = & {\bf F}^{-1}{\bf G} \\ {\bf J}_{p \theta} & = &
{\bf F}^{-1}{\bf H}{\bf J}_{\theta \theta} \\ {\bf F} &=& [{\bf
w}_1 {\bf w}_2 {\bf w}_3]^T \\ {\bf G} &=& \mbox{diag} ({\bf G}_i)
\\ {\bf G}_i & = & \mat{ccc}{1 & {\bf w}_i^T {\bf R}_i & \rho_i ({\bf
R}_i {\bf e}_1 \times {\bf w}_i )^T {\bf R}_i}
\\
{\bf H} &=& - \mat{ccc}{{\bf R}_1 {\bf c}_0 \times {\bf w}_1 &
{\bf R}_2 {\bf c}_0 \times {\bf w}_2 & {\bf R}_3 {\bf c}_0 \times
{\bf w}_3} \\ {\gbf \epsilon}_{p} &=& ({\gbf \epsilon}_{p1}^T,
{\gbf \epsilon}_{p2}^T, {\gbf \epsilon}_{p3}^T)^T \\ {\gbf
\epsilon}_{pi} & = & (\delta L_i, \delta {\bf e}_i^T, \delta
\theta_{Ai}^T)^T
\end{eqnarray}

$\delta L_i = (\delta L_{i1} + \delta L_{i2})/2$ is the mean value
of the variations in links $\overline{B_{i1}C_{i1}}$ and
$\overline{ B_{i2}C_{i2}}$, {\it i.e.}: the variation in the
length of the $i^{th}$ parallelogram. ${\gbf \epsilon}_{p}$ is the
set of the variations in design parameters, which should be
responsible for the position errors of the end-effector, except
the ones which should be responsible for its orientation errors,
{\it i.e.}: ${\gbf \epsilon}_{\theta}$. ${\gbf \epsilon}_{p}$ is
made up of three kinds of errors: the variation in the length of
the $i^{th}$ parallelogram, {\it i.e.}: $\delta L_i, i=1,2,3$, the
position errors of points $A_i$, $B_i$, and $C_i$, {\it i.e.}:
$\delta {\bf e}_i, i=1,2,3$, and the orientation errors of the
directions of the prismatic joints, {\it i.e.}: $\delta
\theta_{Ai}$, $i=1,2,3$. Besides, ${\bf F}$ is nonsingular and its
inverse ${\bf F}^{-1}$ exists because ${\bf F}$ corresponds to the
Jacobian kinematic matrix of the manipulator, which is not
singular when $P$ covers $C_u$, \cite{CHABLAT03}.

According to eq.(\ref{eq:position-error-mapping-function}) and as
$({\bf R}_i {\bf e}_1 \times {\bf w}_i)^T \bot {\bf R}_i {\bf
e}_1$, matrix ${\bf J}_{pp}$ can be simplified by eliminating its
columns associated with $\delta \theta_{Aix}, \ i=1,2,3$. Finally,
thirty-three variations: $\delta L_i$, $\delta e_{ix}$, $\delta
e_{iy}$, $\delta e_{iz}$, $\delta \theta_{Aix}$, $\delta
\theta_{Aiy}$, $\delta \theta_{Aiz}$, $\delta l_i$, $\delta m_i$,
$\delta \gamma_{ix}$, $\delta \gamma_{iy}$, $i=1,2,3$, should be
responsible for the position error of the end-effector.

Rearranging matices ${\bf J}_{pp}$ and ${\bf J}_{p \theta}$, the
position error of the end-effector can be expressed as:
\begin{equation}\label{delra-r2}
  \delta{\bf p} = {\bf J} \ {\gbf \epsilon}_q = \mat{ccc}{{\bf J}_1 & {\bf J}_2 & {\bf J}_3}({\gbf \epsilon}_{q1} \ {\gbf \epsilon}_{q2}
\ {\gbf \epsilon}_{q3})^T
\end{equation}

\noindent with ${\gbf \epsilon}_{qi} = \big(\delta L_i, \delta
e_{ix}, \delta e_{iy}, \delta e_{iz}, \delta \theta_{Aix}, \delta
\theta_{Aiy}, \delta \theta_{Aiz}, \delta l_i, \delta m_i, \delta
\gamma_{ix},
\delta \gamma_{iy} \big)$, and ${\bf J} \in {\field{R}}^{3 \times 33}$.\\

{\bf Sensitivity Indices:} In order to investigate the influence
of the design parameters errors on the position and the
orientation of the end-effector, sensitivity indices are required.
According to section
\ref{section:linkage-kinematic-analysis-results}, variations in
the design parameters of the same type from one leg to the other
have the same influence on the location of the end-effector. Thus,
assuming that variations in the design parameters are independent,
the sensitivity of the position of the end-effector to the
variations in the $k^{th}$ design parameter responsible for its
position error, {\it i.e.}: ${\epsilon}_{q(1,2,3)k}$, is called
${\mu}_{k}$ and is defined by eq.(\ref{eq:sensi-pos}).

\begin{equation}\label{eq:sensi-pos}
  {\mu}_{k} = \sqrt{\sum_{i=1}^{3} \sum_{m=1}^{3}J_{imk}^{2}} \ ,
  \ k = 1, \cdots, 11
\end{equation}

Likewise, $\nu_r$ is a sensitivity index of the orientation of the
end-effector to the variations in the $r^{th}$ design parameter
responsible for its orientation error, {\it i.e.}:
${\epsilon}_{\theta (1,2,3) r}$. $\nu_r$ follows from
eq.(\ref{eq:expression-Jtheta-theta}) and is defined by
eq.(\ref{eq:sensi-orientation}).


\begin{equation}\label{eq:sensi-orientation}
 {\nu}_{r} = \arccos{\frac{\mathrm{tr}({\bf Q}_r) - 1}{2}}
\end{equation}

\noindent where ${\bf Q}_r$ is the rotation matrix corresponding
to the orientation error of the end-effector, and ${\nu}_{r}$ is a
linear invariant: its global rotation \cite{ANGELES02}.

\begin{equation}\label{eq:matrice-Qr-2}
  {\bf Q}_r = \mat{ccc}{ C_{{\nu}_{zr}} C_{{\nu}_{yr}} & \big( C_{{\nu}_{zr}} S_{{\nu}_{yr}} S_{{\nu}_{xr}} - S_{{\nu}_{zr}} C_{{\nu}_{xr}} \big) & \big( C_{{\nu}_{zr}} S_{{\nu}_{yr}} C_{{\nu}_{xr}} + S_{{\nu}_{zr}} S_{{\nu}_{xr}} \big) \\
  S_{{\nu}_{zr}} C_{{\nu}_{yr}} & \big( S_{{\nu}_{zr}} S_{{\nu}_{yr}} S_{{\nu}_{xr}} + C_{{\nu}_{zr}} C_{{\nu}_{xr}} \big) & \big(S_{{\nu}_{zr}} S_{{\nu}_{yr}} C_{{\nu}_{xr}} - C_{{\nu}_{zr}} S_{{\nu}_{xr}} \big) \\
  -S_{{\nu}_{yr}} & C_{{\nu}_{yr}} S_{{\nu}_{xr}} & C_{{\nu}_{yr}} C_{{\nu}_{xr}}}
\end{equation}

\noindent such that $C_{{\nu}_{xr}}=\cos{{\nu}_{xr}}$,
$S_{{\nu}_{xr}}=\sin{{\nu}_{xr}}$, $C_{{\nu}_{yr}}=
\cos{{\nu}_{yr}}$, $S_{{\nu}_{yr}}~=~\sin{{\nu}_{yr}}$,
$C_{{\nu}_{zr}}=\cos{{\nu}_{zr}}$,
$S_{{\nu}_{zr}}=\sin{{\nu}_{zr}}$, and

\begin{eqnarray}
{\nu}_{xr} = \sqrt{\sum_{j=0}^{2} J_{\theta \theta
  1(6j+r)}^{2}} \ , \ r = 1, \cdots, 6 \\
{\nu}_{yr} = \sqrt{\sum_{j=0}^{2} J_{\theta \theta
  2(6j+r)}^{2}} \ , \ r = 1, \cdots, 6 \\
{\nu}_{zr} = \sqrt{\sum_{j=0}^{2} J_{\theta \theta
  3(6j+r)}^{2}} \ , \ r = 1, \cdots, 6
\end{eqnarray}

Finally, ${\mu}_{k}$ can be employed as a sensitivity index of the
position of the end-effector to the $k^{th}$ design parameter
responsible for the position error. Likewise, ${\nu}_{r}$ can be
employed as a sensitivity index of the orientation of the
end-effector to the $r^{th}$ design parameter responsible for the
orientation error. It is noteworthy that these sensitivity indices
depend on the location of the end-effector.

\subsubsection{Results of the Differential Vector Method}
\label{section:differential-vector-analysis-results}
The sensitivity indices defined by eqs.(\ref{eq:sensi-pos}) and
(\ref{eq:sensi-orientation}) are used to evaluate the sensitivity
of the position and orientation of the end-effector to variations
in design parameters, particularly to variations in the
parallelograms.

\begin{figure}[!h]
\includegraphics[width=44mm]{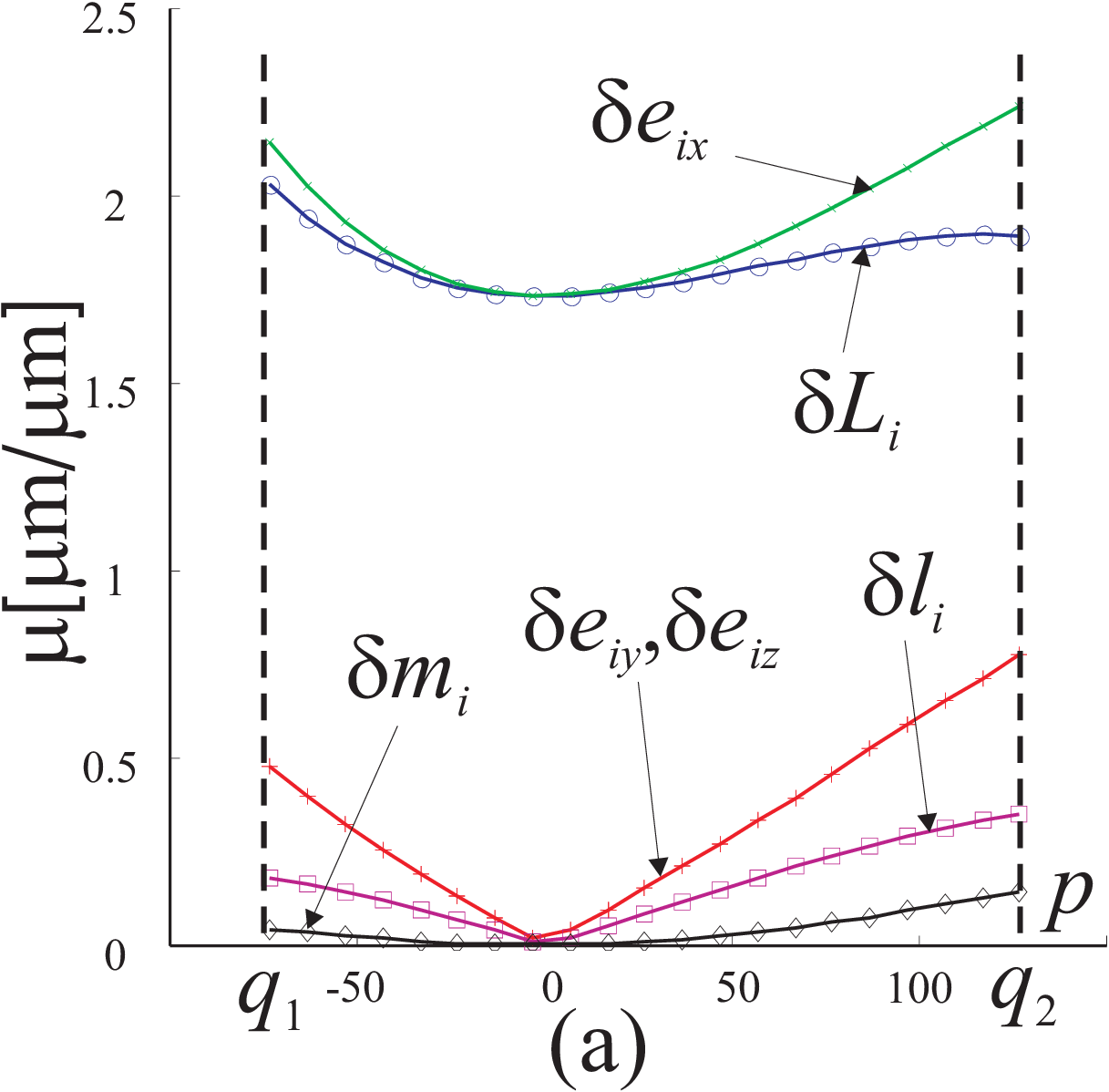} \hfill
\includegraphics[width=44mm]{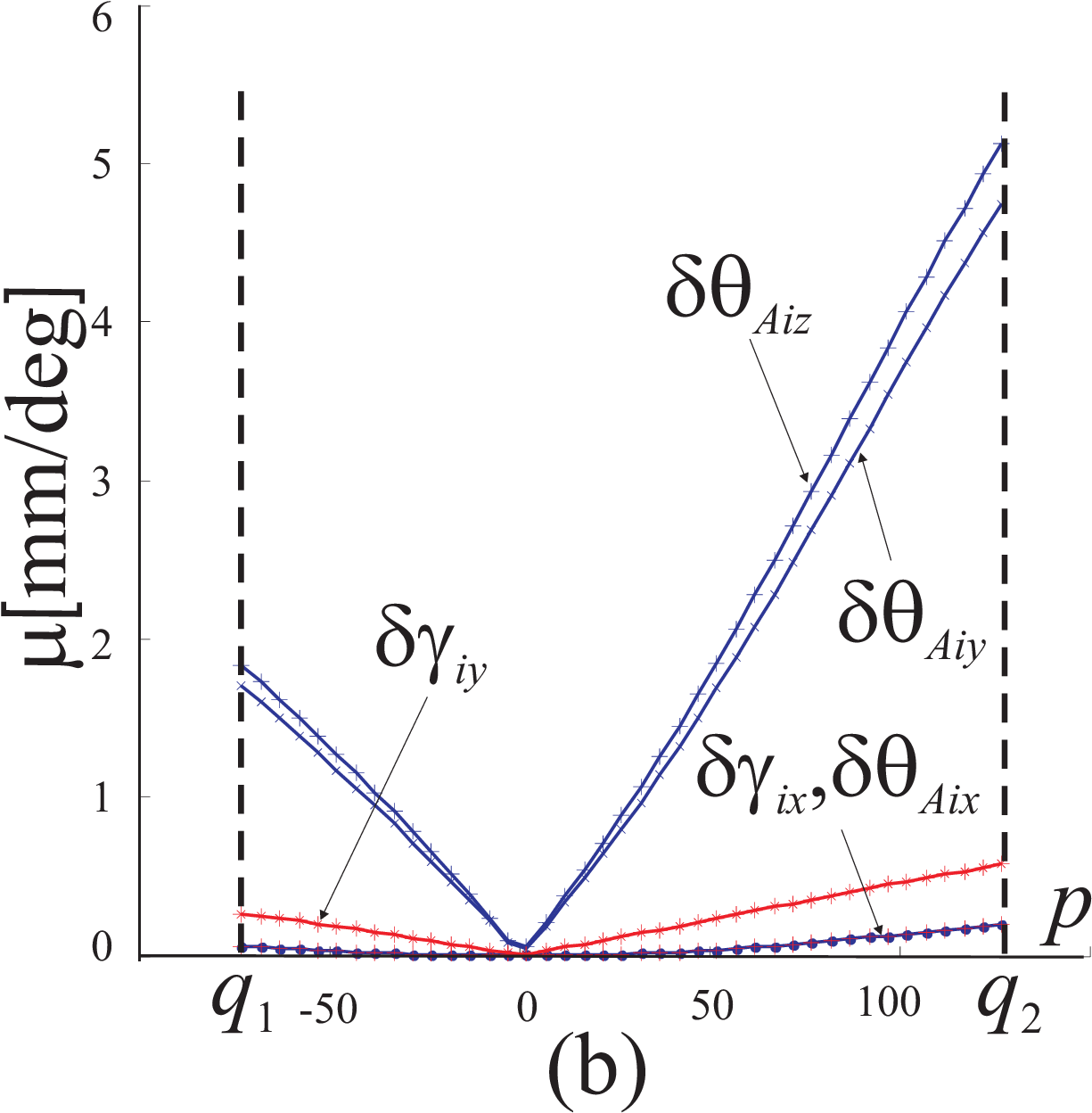}
\caption{Sensitivity of the position of the end-effector along
$Q_1Q_2$, (a): to dimensional variations, (b): to angular
variations} \label{fig:deltar-vardim-varang}
\end{figure}


Figures \ref{fig:deltar-vardim-varang}(a-b) depict the sensitivity
of the position of the end-effector along the diagonal $Q_1Q_2$ of
$C_u$, to dimensional variations and angular variations,
respectively. According to Fig.\ref{fig:deltar-vardim-varang}(a),
the position of the end-effector is very sensitive to variations
in the lengths of the parallelograms, $\delta L_i$, and to the
position errors of points $A_i$, $B_i$, and $C_i$ along axis $x_i$
of ${\cal R}_i$, {\it i.e}: $\delta e_{ix}$. Conversely, the
influence of $\delta l_i$ and $\delta m_i$, the parallelism errors
of the parallelograms, is low and even negligible in the kinematic
isotropic configuration. According to
Fig.\ref{fig:deltar-vardim-varang}(b), the orientation errors of
the prismatic joints depicted by $\delta \theta_{Aiy}$ and $\delta
\theta_{Aiz}$ are the most influential angular errors on the
position of the end-effector. Besides, the position of the
end-effector is not sensitive to angular variations in the
isotropic configuration.

\begin{figure}[!h]
\includegraphics[width=44mm]{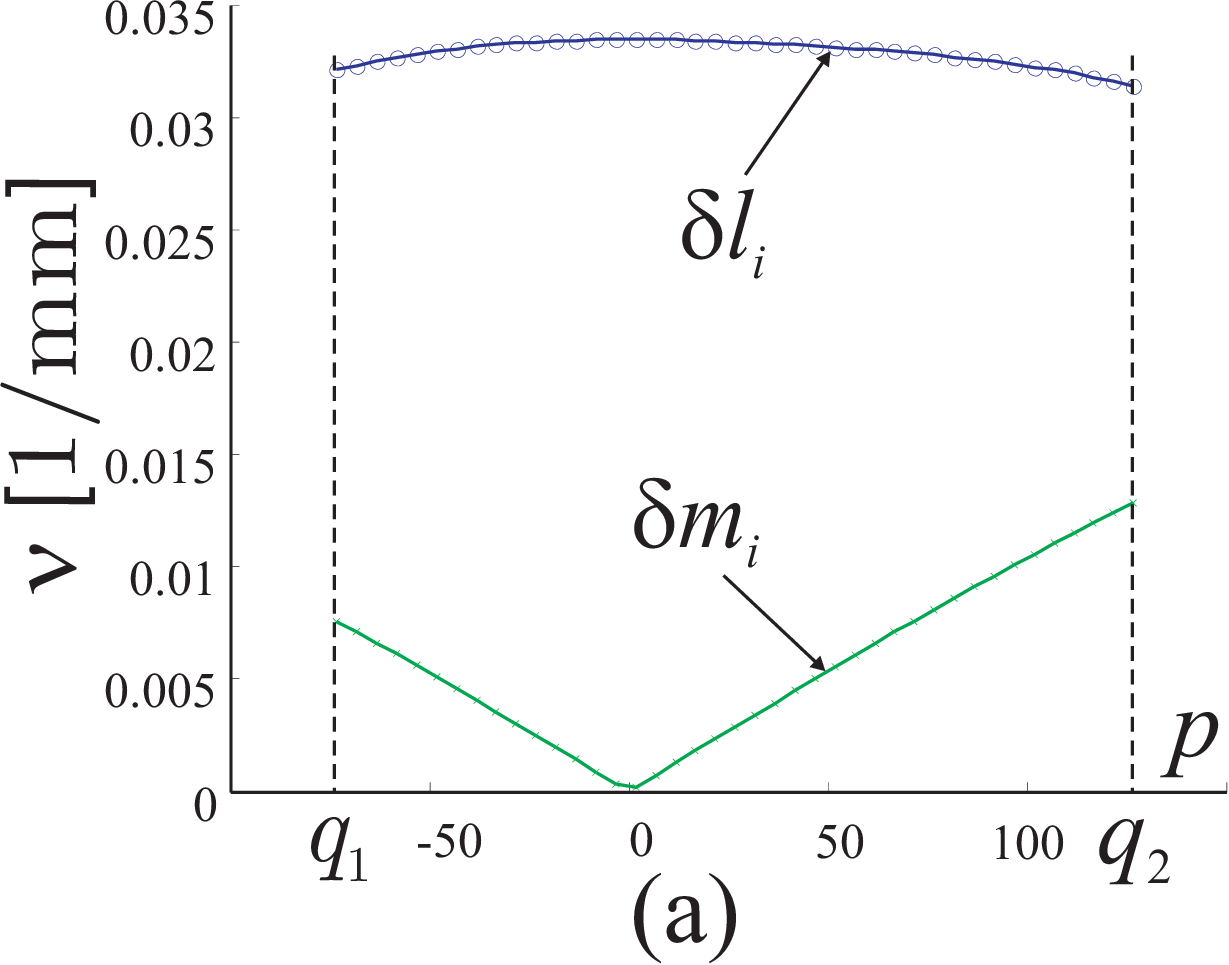} \hfill
\includegraphics[width=44mm]{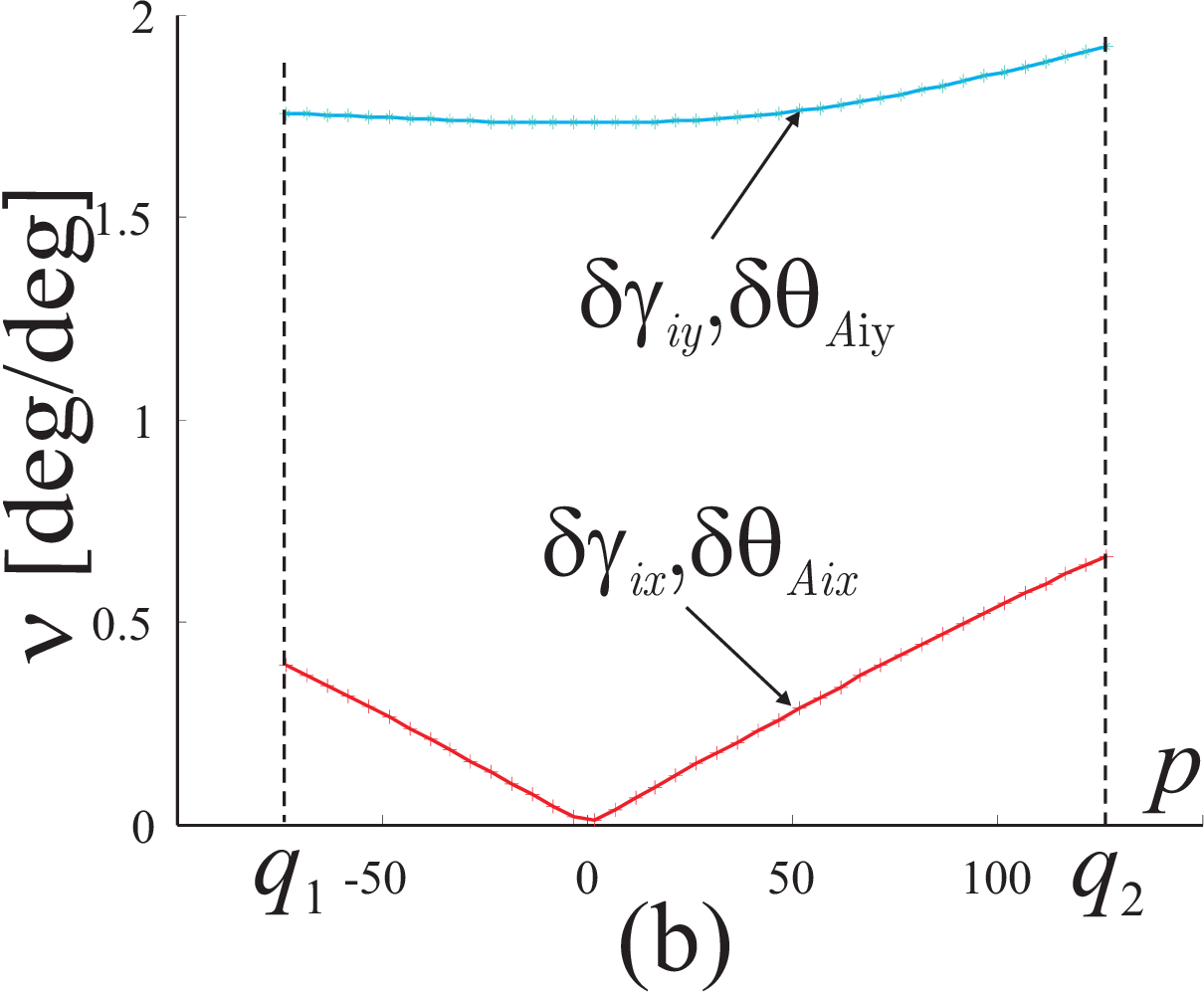}
\caption{Sensitivity of the orientation of the end-effector along
$Q_1Q_2$, (a): to dimensional variations, (b): to angular
variations} \label{fig:deltatheta-vardim-varang}
\end{figure}

Figures \ref{fig:deltatheta-vardim-varang}(a-b) depict the
sensitivity of the orientation of the end effector, along
$Q_1Q_2$, to dimensional and angular variations. According to
Fig.\ref{fig:deltatheta-vardim-varang}(a), $\delta l_i$ and
$\delta m_i$ are the only dimensional variations, which are
responsible for the orientation error of the end-effector.
However, the influence of the parallelism error of the small sides
of the parallelograms, depicted by $\delta l_i$, is more important
than the one of the parallelism error of their long sides,
depicted by $\delta m_i$.

According to figures \ref{fig:deltar-vardim-varang} and
\ref{fig:deltatheta-vardim-varang}, the sensitivity of the
position and the orientation of the end-effector is generally null
in the kinematic isotropic configuration ($p = 0$), and is a
maximum when the manipulator is close to a kinematic singular
configuration, {\it i.e.}: $P \equiv Q_2$. Indeed, only two kinds
of design parameters variations are responsible for the variations
in the position of the end-effector in the isotropic
configuration: $\delta L_i$ and $\delta e_{ix}$. Likewise, two
kinds of variations are responsible for the variations in its
orientation in this configuration: $\delta l_i$, the parallelism
error of the small sides of the parallelograms, $\delta
\theta_{Aiy}$ and $\delta \gamma_{iy}$. Moreover, the
sensitivities of the pose (position and orientation) of the
end-effector to these variations are a minimum in this
configuration, except for $\delta l_i$. On the contrary, $Q_2$
configuration, {\it i.e.}: $P \equiv Q_2$, is the most sensitive
configuration of the manipulator to variations in its design
parameters. Indeed, as depicted by
Figs.\ref{fig:deltar-vardim-varang} and
\ref{fig:deltatheta-vardim-varang}, variations in the pose of the
end-effector depend on all the design parameters variations and
are a maximum in this configuration.

Moreover, figures \ref{fig:deltar-vardim-varang} and
\ref{fig:deltatheta-vardim-varang} can be used to compute the
variations in the position and the orientation of the end-effector
with knowledge of the amount of variations in design parameters.
For instance, let us assume that the parallelism error of the
small sides of the parallelograms, $\delta l_i$, is equal to $10
\mu$m. According to Fig.\ref{fig:deltatheta-vardim-varang}(a), the
position error of the end-effector will be equal about to $3\mu$m
in $Q_1$ configuration ($P \equiv Q_1$). Likewise, according to
Fig.\ref{fig:deltar-vardim-varang}(b), if the orientation error of
the direction of the $i^{th}$ prismatic joint round axis $y_i$ of
${\cal R}_i$ is equal to 1 degree, {\it i.e.}: $\delta
\theta_{Aiy} = 1$~degree, the position error of the end-effector
will be equal about to 4.8~mm in $Q_2$ configuration.

\begin{figure}[!h]
\includegraphics[width=44mm]{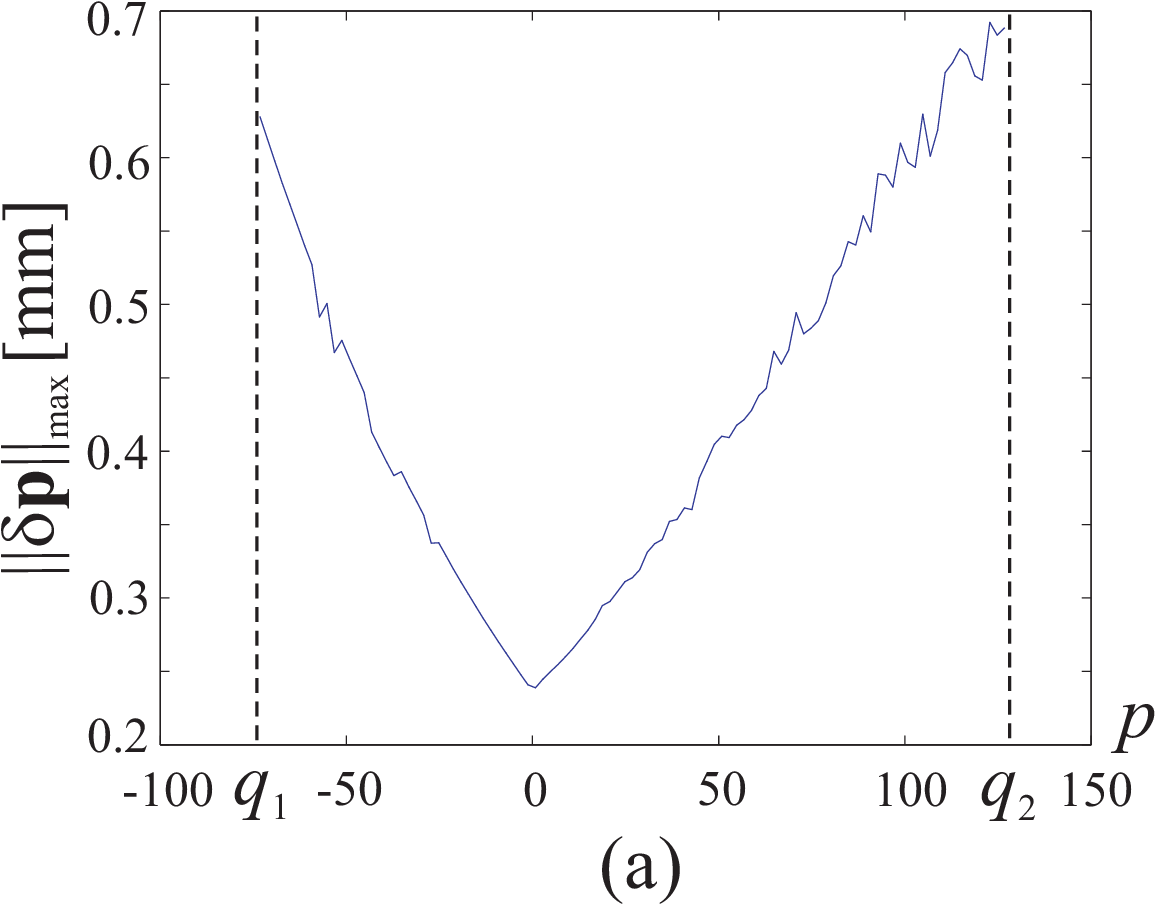} \hfill
\includegraphics[width=44mm]{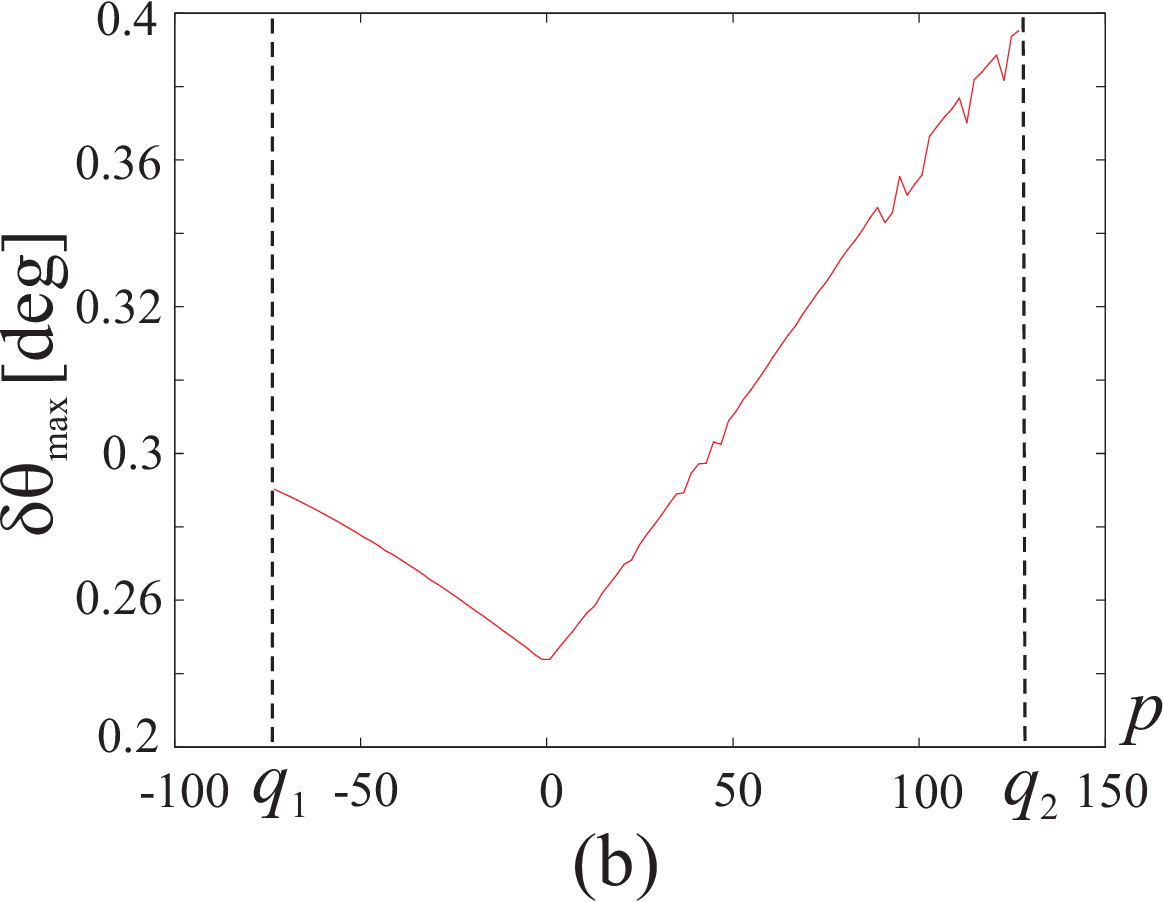}
\caption{Maximum position (a) and orientation (b) errors of the
end-effector along $Q_1Q_2$} \label{fig:calcul-deltap-deltatheta}
\end{figure}

Let us assume now that the length and angular tolerances are equal
to 0.05~mm and 0.03~deg, respectively. Figure
\ref{fig:calcul-deltap-deltatheta}(a) shows the maximum position
error of the end effector when it follows diagonal $Q_1Q_2$ of
cube $C_u$. Likewise, Fig.\ref{fig:calcul-deltap-deltatheta}(b)
shows the maximum orientation error of the end effector along
$Q_1Q_2$. On both sides, the error is a minimum when the
manipulator is in its kinematic isotropic configuration and is a
maximum in $Q_2$ configuration. Besides, the maximum position and
orientation errors of the end-effector are equal to 0.7~mm and
0.4~deg, respectively. These values correspond to the worst case
scenario, which is unlikely.

Then, in order to take into account realistic machining
tolerances, let us assume that the distribution of length and
angular variations is normal.

\begin{figure}[!h]
\includegraphics[width=44mm]{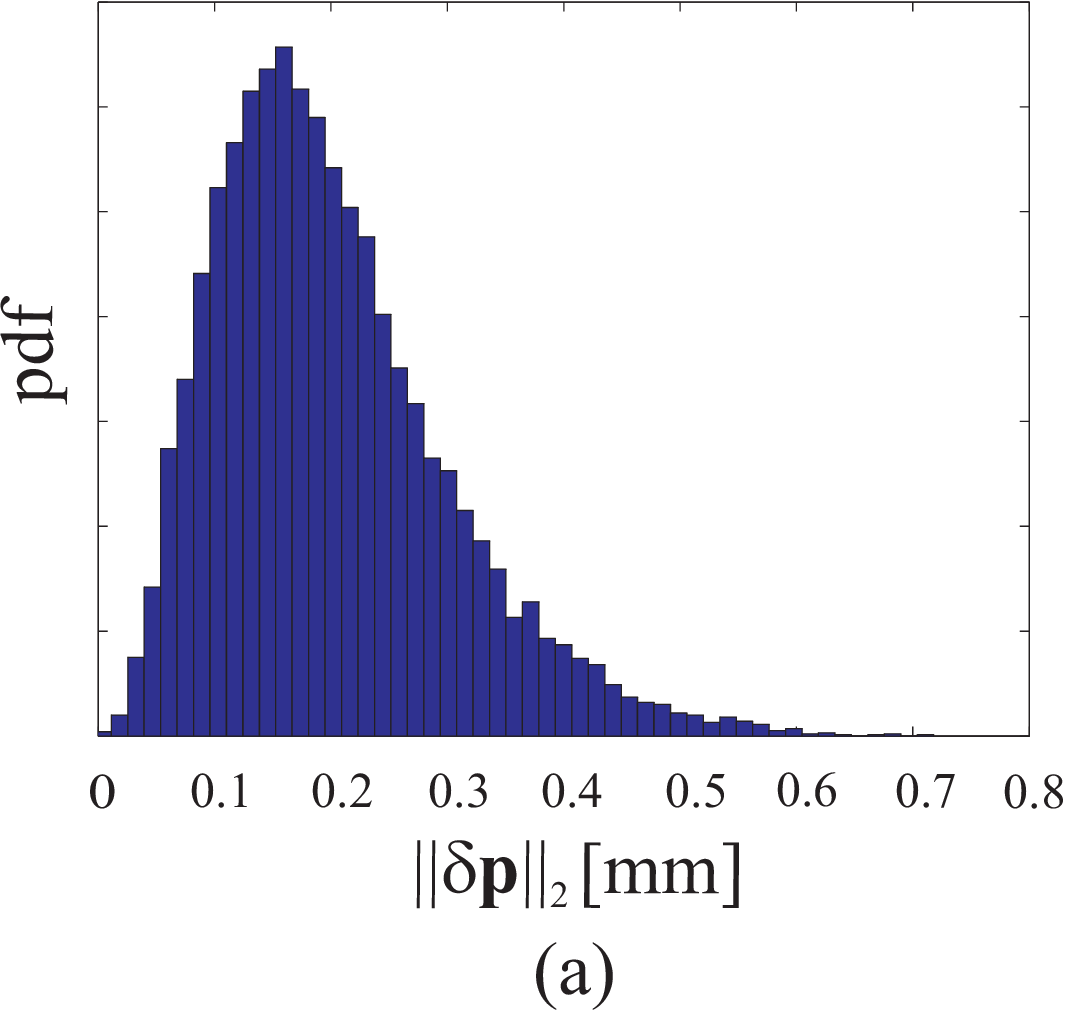} \hfill
\includegraphics[width=44mm]{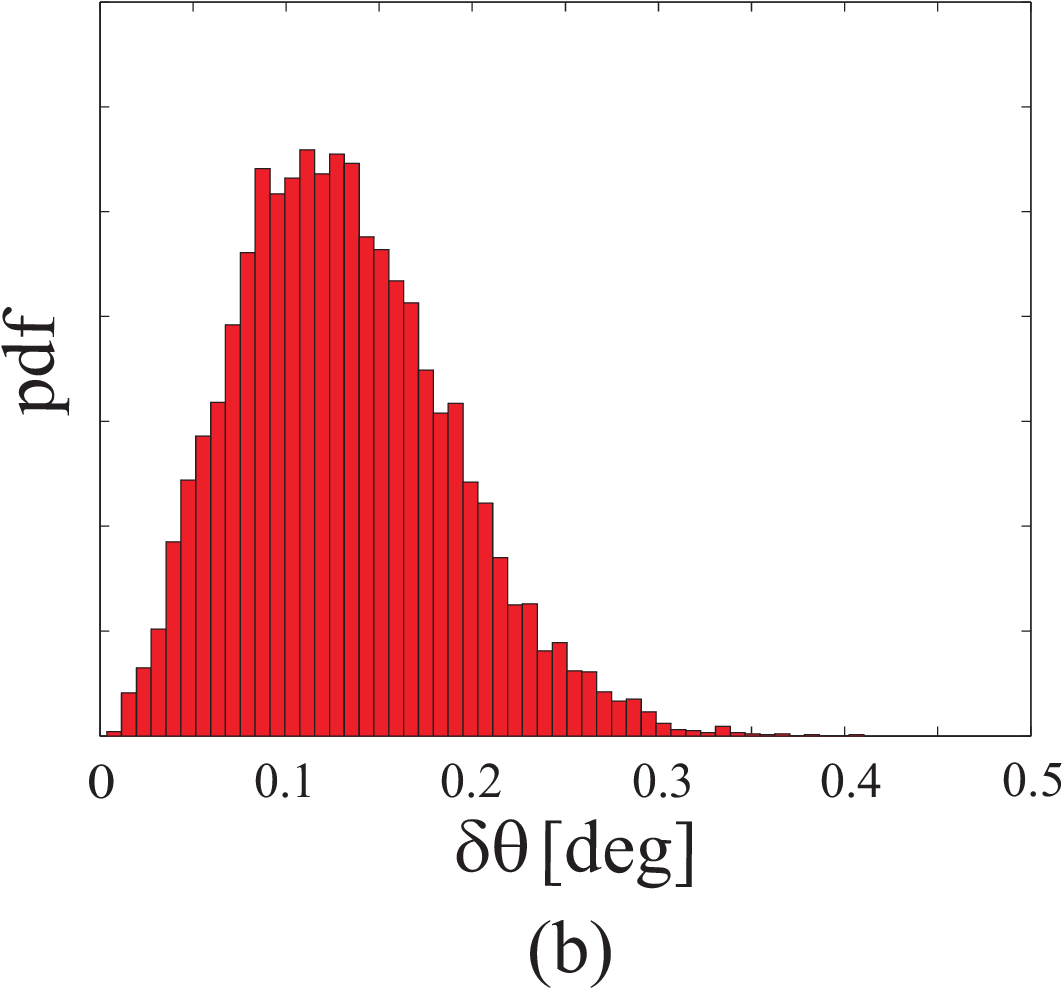}
\caption{Repartition of the position error (a) and the orientation
error (b) of the end-effector in $Q_1$ configuration}
\label{fig:distr-Q1}
\end{figure}

Figures \ref{fig:distr-Q1} (a) and (b) illustrate the repartition
of the position and the orientation errors of the end-effector in
$Q_1$ configuration, respectively.

\begin{figure}[!h]
\includegraphics[width=44mm]{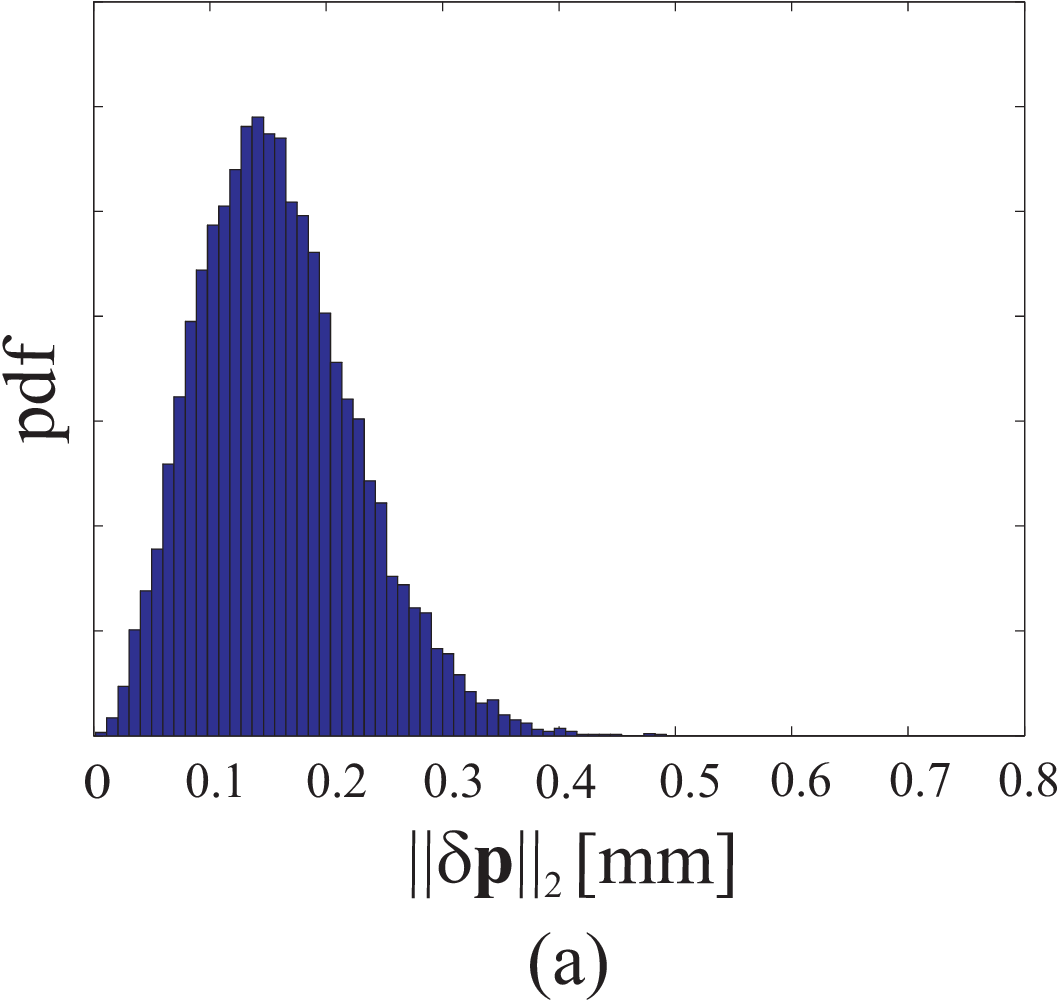} \hfill
\includegraphics[width=44mm]{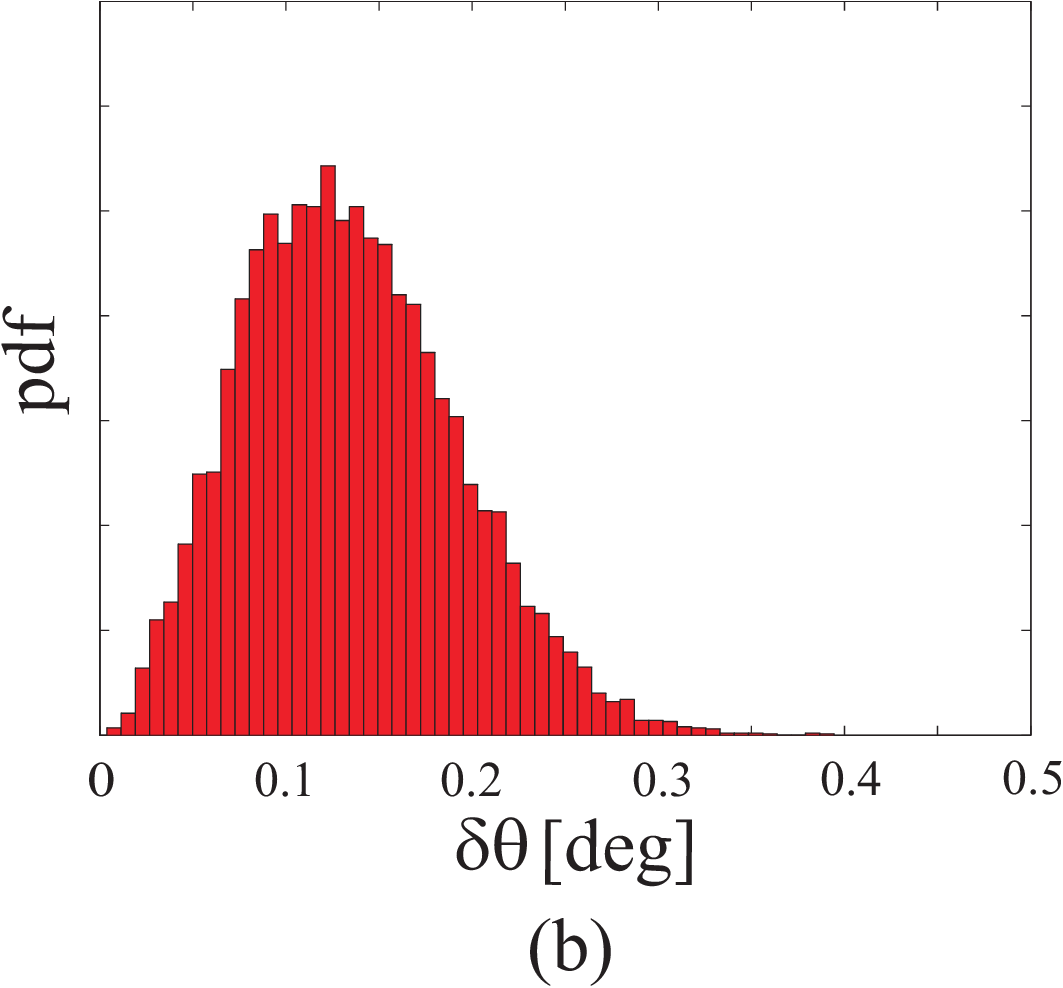}
\caption{Repartition of the position error (a) and the orientation
error (b) of the end-effector in the isotropic configuration}
\label{fig:distr-iso}
\end{figure}

Figures \ref{fig:distr-iso} (a) and (b) illustrate the repartition
of the position and the orientation errors of the end-effector in
the kinematic isotropic configuration, respectively.

\begin{figure}[!h]
\includegraphics[width=44mm]{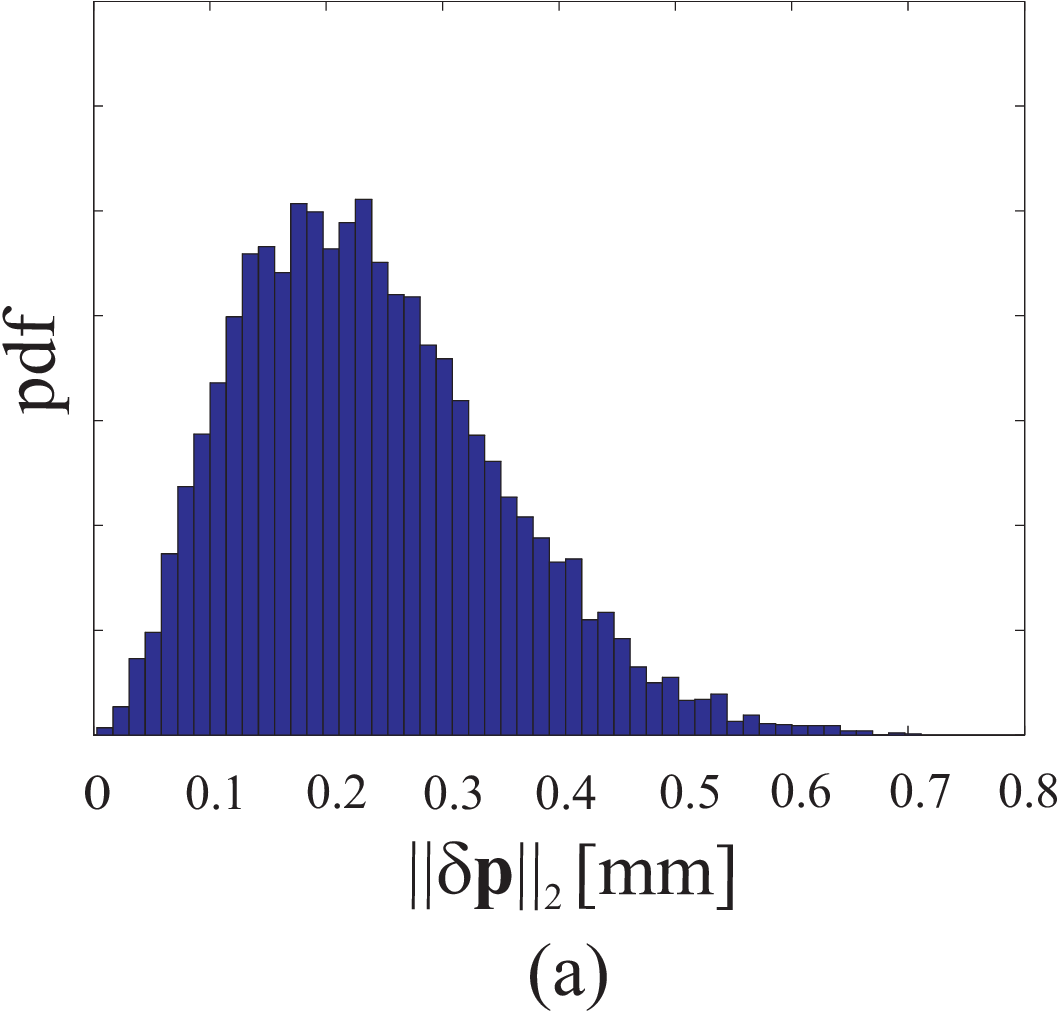} \hfill
\includegraphics[width=44mm]{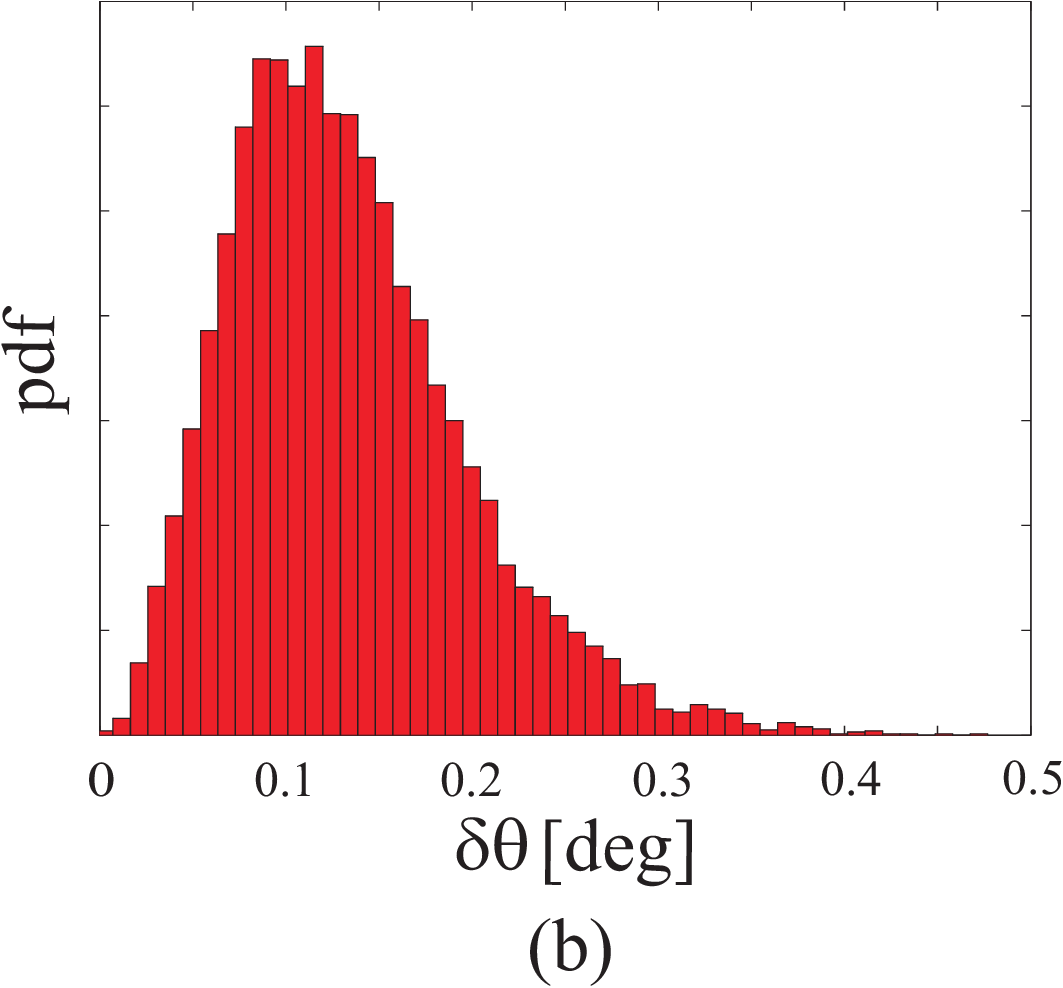}
\caption{Repartition of the position error (a) and the orientation
error (b) of the end-effector in $Q_2$ configuration}
\label{fig:distr-Q2}
\end{figure}

Figures \ref{fig:distr-Q2} (a) and (b) illustrate the repartition
of the position and the orientation errors of the end-effector in
$Q_2$ configuration, respectively.

In Figures \ref{fig:distr-Q1}, \ref{fig:distr-iso},
\ref{fig:distr-Q2} (a) ((b), resp.) , the horizontal axis depicts
the Euclidean norm of the position (orientation, resp.) error of
the end-effector and the vertical axis depicts the corresponding
probability density function (pdf). To plot these figures, we
computed the position and orientation errors of the end-effector
corresponding to more than three thousands sets of geometric
variations following a normal distribution. For example, we can
deduce from these calculations the probabilities to get a position
error lower than 0.3~mm and an orientation error lower than
0.25~deg in $Q_1$, the isotropic, and $Q_2$ configurations.

\begin{table}[!htb]
\centering
\begin{center}
\begin{tabular}{|c|c|c|c|} \cline{2-4}
\multicolumn{1}{c|}{} & \multicolumn{3}{c|}{Configuration} \\
\cline{2-4}
\multicolumn{1}{c|}{} & $Q_1$  & Isotropic & $Q_2$ \\
\hline
  Prob\big($||\delta_p||_2 \leq 0.3$~mm\big) & 0.8468 & 0.9683 & 0.7276 \\
  \hline
  Prob\big($\delta \theta \leq 0.25$~deg\big) & 0.9691 & 0.9690 & 0.9453 \\
  \hline
\end{tabular}
\end{center}
\caption{Probabilities to get a position error lower than 0.3~mm
and an orientation error lower than 0.25~deg in $Q_1$, the
isotropic, and $Q_2$ configurations} \label{tab:probabilities}
\vspace{1mm}
\end{table}

According to Table \ref{tab:probabilities}, the probability to get
a position error lower than 0.3~mm is higher in the kinematic
isotropic configuration than in $Q_1$ and $Q_2$ configurations.
However, the probability to get an orientation error lower than
0.25~deg is the same in $Q_1$ and the isotropic configurations,
but is lower in $Q_2$ configuration.

\section{Conclusions}
In this paper, two complementary methods were introduced to
analyze the sensitivity of a three degree-of-freedom (DOF)
translational Parallel Kinematic Machine (PKM) with orthogonal
linear joints: the Orthoglide. Although these methods were applied
to a particular PKM, they can be readily applied to 3-DOF
Delta-Linear PKM such as ones with their linear joints parallel
instead of orthogonal. Indeed, the input-output equations can be
set in a very similar way since all Delta-linear PKM have
identical leg kinematics, the only difference being in the closure
equations \cite{CHABLAT04}.

On the one hand, a linkage kinematic analysis method was proposed
to have a rough idea of the influence of the length variations of
the manipulator on the location of its end-effector. On the other
hand, a differential vector method was developed to study the
influence of the length and angular variations in the parts of the
manipulator on the position and orientation of its end-effector.
This method has the advantage of taking into account the
variations in the parallelograms.

According to the first method, variations in design parameters of
the same type from one leg to another have the same effect on the
end-effector. Besides, the position of the end-effector is very
sensitive to variations in the lengths of parallelograms and
prismatic joints. The second method shows that the parallelism
errors of the bars of parallelograms are little influential on the
position of the end-effector. Nevertheless, the orientation of the
end-effector of the manipulator is more sensitive to the
parallelism errors of the small sides of the parallelograms than
to the ones of their long sides. Furthermore, it turns out that
the sensitivity of the pose of the end-effector of the manipulator
to geometric variations is a minimum in its kinematic isotropic
configuration. On the contrary, this sensitivity approaches its
maximum close to the kinematic singular configurations of the
manipulator.

Therefore, these results should help the designer of the
Orthoglide to synthesize its dimensional tolerances. Likewise,
these methods can be applied to other Delta-Linear PKM with an aim
of tolerance synthesis. Finally, the next steps in our research
work are to compare the sensitivity of Delta-Linear PKM to
variations in their geometric parameters, and to study the
relation between the sensitivity and the stiffness of such
manipulators.

\begin{acknowledgment}
This research, which is part of {\it ROBEA-MP2} project, was
supported by the CNRS (National Center of Scientific Research in
France).
\end{acknowledgment}


\end{document}